\documentclass[sigconf]{acmart}
\AtBeginDocument{%
  \providecommand\BibTeX{{%
    \normalfont B\kern-0.5em{\scshape i\kern-0.25em b}\kern-0.8em\TeX}}}

\setcopyright{none}
\makeatletter
\renewcommand\@formatdoi[1]{\ignorespaces}
\makeatother
\renewcommand{\footnotetextcopyrightpermission}[1]{}
\acmSubmissionID{5905}

\usepackage{bm}
\usepackage{subfigure}
\usepackage{multirow}
\usepackage{array}

\begin{document}

\title{ShapeSpeak: Body Shape-Aware Textual Alignment for Visible-Infrared Person Re-Identification}
\author{
    Shuanglin Yan\textsuperscript{\rm 1}, Neng Dong\textsuperscript{\rm 1}, Shuang Li\textsuperscript{\rm 2}, Rui Yan\textsuperscript{\rm 1}, Hao Tang\textsuperscript{\rm 3}, Jing Qin\textsuperscript{\rm 3}
}
\affiliation{
    \institution{\textsuperscript{\rm 1}Nanjing University of Science and Technology\\
    \textsuperscript{\rm 2}Chongqing University of Posts and Telecommunications\\
    \textsuperscript{\rm 3}The Hong Kong Polytechnic University}
    \country{}
}





  


\renewcommand{\shortauthors}{Shuanglin Yan, et al.}

\begin{abstract}
Visible-Infrared Person Re-identification (VIReID) aims to match visible and infrared pedestrian images, but the modality differences and the complexity of identity features make it challenging. Existing methods rely solely on identity label supervision, which makes it difficult to fully extract high-level semantic information. Recently, vision-language pre-trained models have been introduced to VIReID, enhancing semantic information modeling by generating textual descriptions. However, such methods do not explicitly model body shape features, which are crucial for cross-modal matching. To address this, we propose an effective Body Shape-aware Textual Alignment (\textbf{BSaTa}) framework that explicitly models and utilizes body shape information to improve VIReID performance. Specifically, we design a Body Shape Textual Alignment (BSTA) module that extracts body shape information using a human parsing model and converts it into structured text representations via CLIP. We also design a Text-Visual Consistency Regularizer (TVCR) to ensure alignment between body shape textual representations and visual body shape features. Furthermore, we introduce a Shape-aware Representation Learning (SRL) mechanism that combines Multi-text Supervision and Distribution Consistency Constraints to guide the visual encoder to learn modality-invariant and discriminative identity features, thus enhancing modality invariance. Experimental results demonstrate that our method achieves superior performance on the SYSU-MM01 and RegDB datasets, validating its effectiveness.
\end{abstract}



\maketitle


\section{Introduction}
Person re-identification (ReID) aims to retrieve images of the same individual across different, non-overlapping cameras. While visible-spectrum ReID~\cite{AIESL, pifhsa, cdvr, li2023logical} has made remarkable progress, it still struggles under low-light or nighttime conditions due to the inability of visible cameras to capture discriminative features. To address this, visible-infrared person re-identification (VIReID)~\cite{GLMC} has emerged. Infrared cameras can capture person appearance regardless of lighting conditions, enhancing the applicability of ReID. However, the main challenge of VIReID lies in the substantial modality gap between infrared and visible images, which complicates cross-modal feature alignment and identity matching, presenting a major obstacle to cross-modal ReID systems.

To reduce the significant modality gap caused by heterogeneous data sources, existing methods are broadly categorized into generative and non-generative approaches. Generative methods~\cite{AlignGAN, msa, acd}, especially those based on generative adversarial networks (GANs)~\cite{GAN}, attempt to synthesize cross-modal images or intermediate domain representations to facilitate learning. These methods provide direct pixel-level supervision, reducing the domain gap at the image level. However, generated images often contain artifacts and inconsistencies, which can mislead the model. Additionally, the instability of generative training and the need for careful tuning make these approaches challenging to scale and deploy in real-world scenarios. In contrast, non-generative methods~\cite{DCLNet, CAJ+, rle} focus on learning a shared feature space for directly aligning infrared and visible representations. This is typically achieved through network design, metric learning strategies, or contrastive objectives that enforce cross-modal similarity. Although these methods avoid the instability of image synthesis, their exclusive reliance on identity-level supervision constrains the model’s capacity to capture high-level semantics, making it challenging to maintain both modality-invariant and identity-discriminative features simultaneously.

In recent years, vision-language pretraining (VLP) models have rapidly advanced, leveraging language supervision to train visual encoders with exceptional high-level semantic extraction capabilities. Inspired by this, study~\cite{csdn} was the first to introduce the VLP model CLIP~\cite{clip} to address the aforementioned challenges. This approach generates separate textual descriptions for visible and infrared images, and integrates their textual representations to guide the visual network in learning identity-related high-level semantic information. Specifically, it employs a vision-language alignment mechanism to introduce semantic supervision in VIReID, mitigating modality mismatches caused by significant differences in appearance, lighting conditions, and texture between infrared and visible images. The core assumption is that textual descriptions can provide additional high-level semantic information, such as body shape and clothing categories, thereby compensating for the modality misalignment that may arise when relying solely on image-level features. However, this method still has notable limitations. 
While the method attempts to learn body shape information through textual descriptions, it does not explicitly model body shape features, instead relying on the model to automatically extract such information during contrastive learning. Since body shape is inherently modality-invariant and critical for VIReID, relying solely on implicit learning may hinder the model's ability to fully capture and leverage body shape features, ultimately limiting recognition performance.

Building upon this, we propose an effective Body Shape-aware Textual Alignment (\textbf{BSaTa}) framework to explicitly extract and utilize body shape information for enhancing feature representation in VIReID. The framework consists of two main components: a Body Shape Textual Alignment (BSTA) module and a Shape-aware Representation Learning (SARL) module. The BSTA module uses an advanced human parsing model to analyze both visible and infrared images, extracting key body shape cues (such as body contours, limb proportions, and torso structure) to generate body shape maps, which are then encoded into visual shape features. These features are transformed into structured textual shape representations using CLIP’s vision-language alignment capabilities. To ensure high-quality textual shape representations, we introduce a Text-Visual Consistency Regularizer (TVCR) that enforces symmetric alignment between the visual and textual shape features in a shared embedding space, enabling the textual shape representations to more accurately describe identity-related shape characteristics. The SARL module incorporates a multi-text supervision mechanism, which not only effectively guides the visual encoder to learn discriminative shape features but also enables the fully extraction of modality-invariant identity information. Furthermore, we design a distribution consistency constraint (DCC) that treats the identity-level text representation as an alignment anchor to reduce the distribution gap between visible and infrared modalities, thus enhancing cross-modal alignment. Compared to existing methods, our method explicitly models body shape information, enabling more effective extraction of modality-shared features and significantly improving recognition accuracy and generalization in VIReID. 

Here are the main contributions of our paper: (1) We propose a body shape-aware textual alignment framework that explicitly extracts and utilizes body shape information, enhancing the model’s ability to capture modality-invariant features. (2) We design a body shape textual alignment module with a cross-modal consistency regularizer that transforms body shape maps into structured textual representations, ensuring the quality of the generated textual representations. (3) We introduce a shape-aware representation learning mechanism that combines multi-text supervision and distribution consistency constraints to emphasize body shape information and enhance feature representation. (4) Extensive experiments demonstrate that our method achieves superior performance on both SYSU-MM01 and RegDB, validating its effectiveness.

\vspace{-0.21cm}
\section{Related Work}
\subsection{Visible-Infrared Person Re-Identification}
Visible-infrared person re-identification (VIReID) faces significant challenges due to the inherent modality gap~\cite{lin2022learning} between infrared and visible images, which arises from their fundamentally different imaging principles. Unlike visible images~\cite{mvip, etndnet, zhang2022cross} that capture detailed textures and colors under natural lighting, infrared images represent thermal radiation, often lacking fine-grained appearance details~\cite{MANet, LCR2S, m3net}. This discrepancy leads to substantial feature misalignment, making cross-modal retrieval highly challenging. Over the years, various strategies have been proposed to bridge this gap, which can be broadly categorized into generative and non-generative approaches.

Generative methods~\cite{AlignGAN, msa, TSME, acd, agpi2} aims to bridge the modality gap by synthesizing cross-modal images or intermediate domain representations, often utilizing generative adversarial networks (GANs)~\cite{GAN} or variational frameworks. For instance, Wang et al.~\cite{AlignGAN} introduced a pixel alignment module based on GANs that transforms real visible images into synthetic infrared-like images to mitigate the cross-modal discrepancy. While these methods help reduce the modality gap, they are often plagued by issues such as semantic distortion, mode collapse, and training instability, which hinder their generalization to real-world scenarios. Non-generative methods~\cite{MSO, DCLNet, MAUM, CAJ+, rle, EEES, csdn} have emerged as the dominant solution for VIReID due to their stability and effectiveness. These approaches~\cite{PMT, ScRL} primarily focus on learning modality-invariant and discriminative features through network architecture design, metric learning strategies, or contrastive objectives that enforce cross-modal similarity. Recently, vision-language pre-trained (VLP) models, such as CLIP~\cite{clip}, have been incorporated into VIReID, providing richer semantic supervision by generating textual descriptions~\cite{csdn}. However, this method mainly emphasizes appearance attributes and overlook the importance of body shape modeling~\cite{wang2022body}, which is equally crucial for cross-modal identity discrimination. This paper addresses this gap by explicitly encoding shape information using structured textual shape representations.

\begin{figure*}[t!]
  \centering
  \setlength{\abovecaptionskip}{0.1cm}
  \setlength{\belowcaptionskip}{-0.2cm}
  \includegraphics[width=\linewidth]{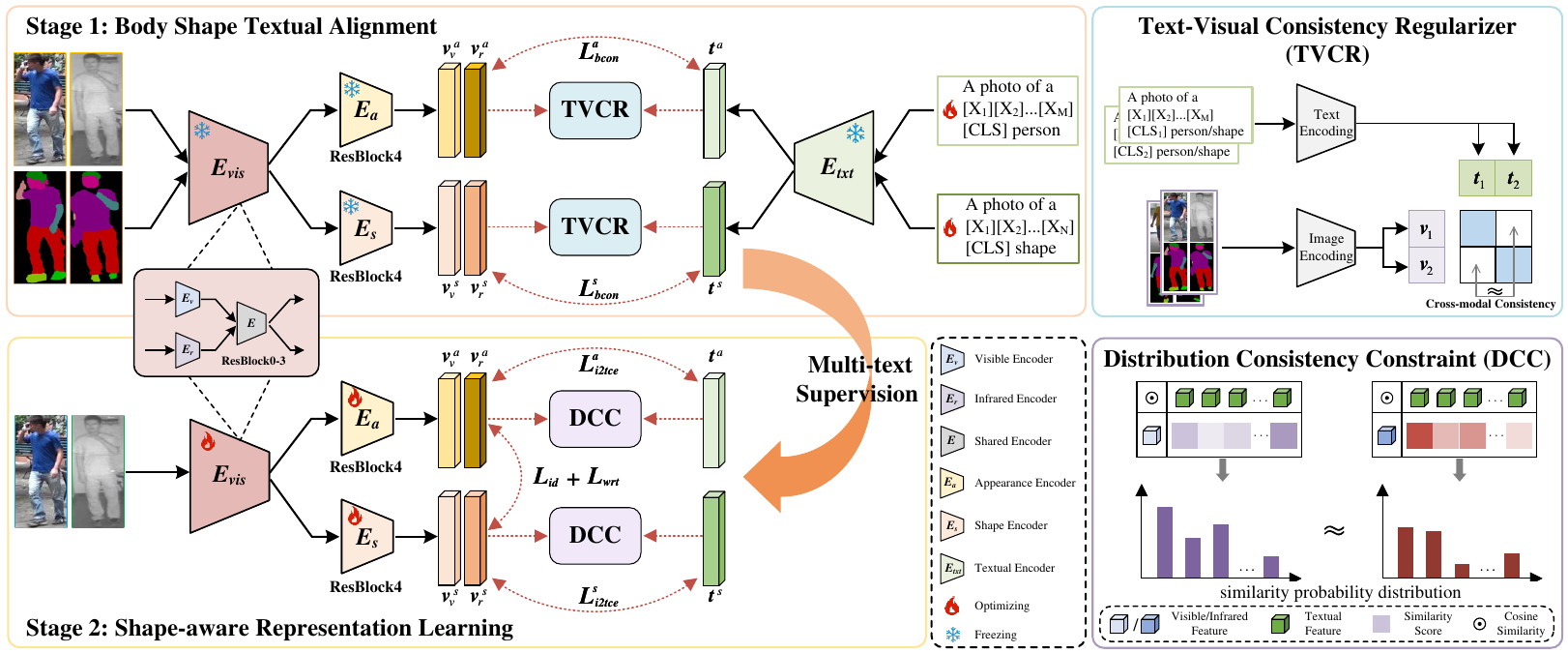}\\
  \caption{Overview of our BSaTa. It consists of two stages: Body Shape Textual Alignment (Stage 1) and Shape-Aware Representation Learning (Stage 2). In Stage 1, we leverage the powerful cross-modal alignment capability of CLIP to generate identity-level textual shape and appearance representations for each infrared-visible image pair. A Text-Visual Consistency Regularizer (TVCR) is further introduced to ensure the semantic quality of the generated textual descriptions. In Stage 2, the textual representations generated in the previous stage are used as supervision to construct a Multi-text Supervision mechanism, which guides the training of the visual encoder. Additionally, a Distribution Consistency Constraint (DCC) is designed to utilize the textual features as anchors to enforce alignment between the feature distributions of infrared and visible modalities, thereby fully mining modality-invariant identity information and enhancing the feature representation capability. During inference, the shape and appearance features of both infrared and visible images are extracted, concatenated, and used for retrieval.}
  \label{Fig:2}
\end{figure*}

\vspace{-4pt}
\subsection{Vision-Language Pre-Training}
The "pre-training and fine-tuning" paradigm has become a cornerstone in computer vision, enabling models to transfer knowledge from large-scale datasets to various downstream tasks. Previous prevailing practice is rooted in the supervised unimodal pre-training~\cite{vit, resnet} on ImageNet~\cite{imagenet}. Initially, pre-training was predominantly conducted on single-modal datasets like ImageNet~\cite{imagenet}, where models such as ResNet~\cite{resnet} and Vision Transformer (ViT)~\cite{vit} relied on category labels for supervised learning to acquire powerful visual representations. While effective, this approach was inherently limited by the need for extensive labeled data and its inability to capture cross-modal relationships. To address these limitations, vision-language pre-training (VLP) has emerged as a promising alternative, shifting the focus from purely visual learning to multimodal representation learning. VLP models aim to bridge the gap between vision and language by leveraging large-scale multimodal data. Early works~\cite{vilbert, visualbert, uniter, albef, beit} explored joint vision-language representations through masked modeling and image-text alignment tasks, demonstrating improved semantic understanding across modalities. However, these methods often relied on complex multimodal fusion architectures, which added computational overhead and limited generalization. More recently, contrastive learning~\cite{clip, align, filip} has gained traction as a scalable and effective alternative. By leveraging contrastive objectives to distinguish between positive and negative pairs, models can learn robust and transferable representations. A notable example is Contrastive Language-Image Pre-training (CLIP)~\cite{clip}, which has shown impressive adaptability across various downstream tasks~\cite{clip4clip, cris, CFine, trimatch}. In this work, we leverage CLIP's powerful semantic understanding to enhance the extraction of identity-related high-level semantics, with the goal of bridging the modality gap in VIReID.

\vspace{-5pt}
\subsection{Prompt Learning}
Prompt learning has emerged as a powerful paradigm for adapting large-scale pre-trained models to specific tasks with minimal modifications. Initially introduced in natural language processing (NLP)~\cite{autoprompt, how}, this approach enables models to conditionally interpret tasks by providing carefully designed prompts. While early methods relied on manually crafted templates, recent advancements have shifted toward automated prompt optimization, mitigating issues such as instability and domain bias. In computer vision, prompt learning has gained traction, particularly in vision-language models. CoOp~\cite{coop} pioneered learnable text prompts for adapting CLIP to various visual recognition tasks, eliminating the need for manual prompt engineering. CoCoOp~\cite{cocoop} further advanced this concept by introducing conditional prompt tuning, which enhances generalization across unseen categories. These developments highlight the potential of prompt learning for leveraging vision-language representations in robust model adaptation. Despite these successes, applying prompt learning to fine-grained tasks~\cite{propot}, such as person re-identification, remains challenging. Recent work, such as CLIP-ReID~\cite{clipreid}, has attempted to bridge this gap by generating pedestrian descriptions for identity discrimination. Inspired by this, we propose leveraging prompt learning to generate modality-shared appearance and body shape descriptions for extracting high-level semantics in VIReID.

\vspace{-4pt}
\section{The BSaTa Framework}
Our proposed BSaTa is a conceptually simple yet effective framework, as shown in Figure \ref{Fig:2}. We begin by extracting visual features from both infrared and visible images using a visual backbone network. These features are then passed through the Body Shape Textual Alignment module, which generates structured textual shape representations. Finally, the generated textual shape representations serve as semantic guidance to supervise the backbone network, enabling it to better capture modality-invariant body shape information and thereby enhancing the feature representation capability for VIReID.  

\vspace{-4pt}
\subsection{Feature Extraction}
Given the sample set $D=\{I^{v}, I^{r}, Y\}$, where $I^{v}=\{ I_{i}^{v}\}_{i=1}^{N_{v}}$ represents the visible image set, $I^{r}=\{I_{i}^{r}\}_{i=1}^{N_{r}}$ is the infrared image set, and $\bm Y=\{\bm y_{i}\}_{i=1}^{N_{y}}$ is the label set. Each $\bm y_{i}\in \mathbb{R}^{N_y}$ is a one-hot ground-truth label vector, and $N_y$ is the number of identities. For a visible-infrared image pair ($I_{i}^{v}$, $I_{i}^{r}$), we use ResNet50 \cite{resnet} as the backbone network to extract visual features. To effectively model modality-specific information, we duplicate the shallow layers of ResNet50 to construct two modality-specific encoders, $\bm E_{v}$ and $\bm E_{r}$, which extract low-level features $\bm v^l_{v,i}$ and $\bm v^l_{r,i}$ from visible and infrared images, respectively. The remaining layers of ResNet50 serve as a modality-shared encoder $\bm E$, which maps the modality-specific low-level features into a joint feature space to learn modality-invariant visual representations $\bm v_{v,i}$ and $\bm v_{r,i}$. Previous studies \cite{clipreid, CFine, csdn} have highlighted the effectiveness of CLIP \cite{clip} in addressing the challenges of ReID, demonstrating that its powerful vision-language joint representation significantly enhances cross-modal matching and semantic understanding. To leverage CLIP’s strong capability in high-level semantic comprehension, we initialize our image backbone network using CLIP's visual encoder.

The goal is to align $\bm v_{v,i}$ and $\bm v_{r,i}$ in the joint feature space, which requires the encoder to effectively extract modality-shared high-level semantic information from both infrared and visible images. To achieve this, work~\cite{csdn} employs a vision-language alignment mechanism to generate textual descriptions for infrared and visible images separately, introducing additional semantic supervision to guide the visual encoder in learning key features such as body shape. Although this approach attempts to capture body shape information through textual descriptions, it does not explicitly model body shape features but instead relies on the model to discover them automatically during contrastive learning. However, relying solely on implicit learning may not be sufficient for the model to fully capture and utilize body shape features, thereby limiting its recognition performance. In light of this, we propose BSaTa, which explicitly extracts and utilizes body shape information, thereby enhancing feature representation for VIReID. It primarily consists of a Body Shape Textual Alignment module, equipped with a Text-Visual Consistency Regularizer, and a Shape-aware Representation Learning module.

\vspace{-4pt}
\subsection{Body Shape Textual Alignment} 
Body shape information is inherently modality-invariant and plays a crucial role in VIReID. However, relying solely on identity-level supervision to implicitly learn body shape features from images is highly challenging, as identity information often encompasses multiple appearance factors that may interfere with the extraction of body shape features. To address this issue, a viable approach is to incorporate an advanced human parsing model to obtain body shape maps from images and use them to generate shape features as supervision signals, guiding the network to better capture body shape information~\cite{ScRL}. However, since shape features are typically extracted from the parsed body shape maps, their expressiveness is constrained by the semantic understanding capability of the visual network, making it difficult to directly support higher-level semantic comprehension. In recent years, vision-language pretraining models, such as CLIP, have demonstrated powerful high-level semantic extraction capabilities in multimodal learning. By leveraging text as a supervision signal, these models significantly enhance the ability of visual networks to understand complex semantics. Inspired by this, our goal is to obtain textual representations that describe body shape information and use them as supervision signals to help the network more effectively learn shape features, thereby improving the robustness of cross-modality matching and feature representation. However, acquiring high-quality textual descriptions of body shape remains a key challenge. To tackle this, we design a Body Shape Textual Alignment (BSTA) module, equipped with a Text-Visual Consistency Regularizer (TVCR).

\textbf{Body Shape Textual Alignment (BSTA) Module}. This module leverages CLIP’s strong vision-language alignment capability to transform body shape maps, generated by the human parsing model, into structured textual shape representations. Specifically, given a visible-infrared image pair ($I_{i}^{v}$, $I_{i}^{r}$), we employ the advanced human parsing model SCHP~\cite{schp} to generate the corresponding shape maps $H_{i}^{v}$ and $H_{i}^{r}$, which are then fed into their respective modality-specific encoding networks for feature extraction. To further refine the representation, we introduce an additional shape encoder $\bm E_{s}$, constructed by duplicating the last block of ResNet50. This encoder functions as a modality-shared module, generating the visual shape features $\bm v^s_{v,i}$ and $\bm v^s_{r,i}$.

After obtaining the visual shape features, we further convert them into modality-aligned textual shape representations based on CLIP. To achieve this, inspired by CLIP-ReID~\cite{clipreid}, we introduce a learnable textual shape description $T^s_i$, formatted as follows:
\begin{equation}\
\begin{aligned}
T^s_i="A~~photo~~of~~a~~[\bm X^s_1][\bm X^s_2]...[\bm X^s_M][CLS]~~shape"
\end{aligned},
\end{equation}
where $[X^s_m]$ represents a trainable shape word token, $[CLS]$ denotes the identity label corresponding to ($I_{i}^{v}$, $I_{i}^{r}$), and $M$ is the total number of learnable shape tokens. This learnable text description is adaptively optimized by aligning it with the corresponding visual shape features, enabling it to more accurately capture the key semantic information of the shape features. We then use CLIP's textual encoder $\bm E_{t}$ to extract the textual shape feature $\bm t^s_i$ from $T^s_i$, thereby obtaining a modality-aligned textual shape representation. This textual shape representation $\bm t^s_i$ is identity-level and modality-invariant, remaining unaffected by variations in illumination, viewpoint, and other factors. Essentially, it serves as a shape prototype. Therefore, we align the textual shape representation $\bm t^s_i$ with the visual shape features ($\bm v^s_{v,i}$ and $\bm v^s_{r,i}$) using a bidirectional cross-modal contrastive loss, as follows:
\begin{equation}\
\begin{aligned}
\mathcal{L}^s_{bcon}=\mathcal{L}^s_{i2t}+\mathcal{L}^s_{t2i}
\end{aligned},
\end{equation}
\begin{equation}
\begin{aligned}
\mathcal{L}^s_{i2t} =  &- \frac{1}{{{n_v}}}\sum\limits_{i = 1}^{{n_v}} {log\frac{{exp({sim({\bm v^s_{v,i}, \bm t^s_i})})}}{{\sum\nolimits_{j = 1}^{{n_v}} {exp( {sim({\bm v^s_{v,i}, \bm t^s_j})})} }}}  \\&- \frac{1}{{{n_r}}}\sum\limits_{i = 1}^{{n_r}} {log\frac{{exp( {sim({\bm v^s_{r,i},\bm t^s_i})})}}{{\sum\nolimits_{j = 1}^{{n_r}} {exp({sim({\bm v^s_{r,i}, \bm t^s_j})})} }}} 
\end{aligned},
\label{Eq:3}
\end{equation}
\begin{equation}
	\begin{aligned}
	\mathcal{L}^s_{t2i} =  &- \frac{1}{{{n_v}}}\sum\limits_{i = 1}^{{n_v}} {\frac{1}{{\left| {P\left( {{y_i}} \right)} \right|}}\sum\limits_{{p_i} \in P\left( {{y_i}} \right)} {log\frac{{exp({sim({\bm v^s_{v,p_i},\bm t^s_{y_i}})})}}{{\sum\nolimits_{j = 1}^{{n_v}} {exp({sim({\bm v^s_{v,j}, \bm t^s_{y_i}})})} }}} }\\ 
	&- \frac{1}{{{n_r}}}\sum\limits_{i = 1}^{{n_r}} {\frac{1}{{\left| {P\left( {{y_i}} \right)} \right|}}\sum\limits_{{p_i} \in P\left( {{y_i}} \right)} {log \frac{{exp ( {sim({\bm v^s_{r,p_i},\bm t^s_{y_i}})})}}{{\sum\nolimits_{j = 1}^{{n_r}} {exp ( {sim({\bm v^s_{r,j}, \bm t^s_{y_i}})})} }}} }  \\ 
	\end{aligned},
 \label{Eq:4}
\end{equation}
where $n_{v}=n_{r}$, and $sim(\cdot)$ denotes the cosine similarity function. $\bm t^s_{y_i}$ is the textual shape feature for identity $y_{i}$, while $P(y_{i})$ represents the set of sample indices corresponding to identity $y_{i}$, with $|P(y_{i})|$ denoting the cardinality of $P(y_{i})$. 

\textbf{Text-Visual Consistency Regularizer}. To strengthen text-visual alignment, we introduce a Text-Visual Consistency Regularizer on top of the bidirectional cross-modal contrastive loss, ensuring bidirectional consistency in text-image matching. While the contrastive loss optimizes the relative similarity between positive and negative pairs, it does not explicitly enforce the symmetry of the similarity matrix across modalities, which could lead to unstable matching and inconsistent modality distributions. By incorporating the symmetry constraint, we promote symmetric alignment between images and text in the shared feature space, thereby reducing modality discrepancies. This enables the textual shape representation to more accurately describe the visual shape features and enhances the reliability of the generated textual shape representation. The text-visual consistency regularizer is formulated as follows:
\begin{equation}\
\begin{aligned}\label{Eq:5}
\mathcal{L}^s_{tvcr}=&\frac{1}{n_v}\sum_{i=1}^{n_v}\sum_{j=1}^{n_v}(sim(\bm v^s_{v,i}, \bm t^s_j)-sim(\bm v^s_{v,j}, \bm t^s_i))^2\\
&\frac{1}{n_r}\sum_{i=1}^{n_r}\sum_{j=1}^{n_r}(sim(\bm v^s_{r,i}, \bm t^s_j)-sim(\bm v^s_{r,j}, \bm t^s_i))^2
\end{aligned}.
\end{equation}
In this way, we generate dependable textual shape representations that serve as robust semantic descriptions of visual shape features. These textual representations are then used to guide the training of the visual encoder, enabling it to accurately capture identity-related, modality-invariant body shape information. This effectively reduces the modality gap and enhances cross-modal matching performance. 

Furthermore, to leverage CLIP’s powerful semantic understanding for enhancing the visual encoder’s ability to capture modality-shared appearance features, we follow CSDN~\cite{csdn} and introduce a learnable, modality-invariant, identity-level appearance description $T^a_i="A~~photo~~of~~a~~[\bm X^a_1][\bm X^a_2]...[\bm X^a_N][CLS]~~person"$, where $N$ is the total number of learnable shape tokens. This description is then processed by CLIP’s textual encoder to extract textual appearance representations. We refer to $\bm v_{v,i}$ and $\bm v_{r,i}$ as the visual appearance features $\bm v^a_{v,i}$ and $\bm v^a_{r,i}$, respectively. Subsequently, we guide the learning of the modality-shared text appearance description using a bidirectional cross-modal alignment loss $\mathcal{L}^a_{bcon}$, based on visual appearance features, formulated similarly to Equations (2), (3), and (4). Similarly, the consistency regularizer is applied to the textual appearance representation, denoted as $\mathcal{L}^a_{tvcr}$.

\vspace{-4pt}
\subsection{Shape-aware Representation Learning}
This paper aims to enhance the visual encoder's ability to perceive body shape information, facilitating a more comprehensive extraction of modality-shared features and improving cross-modal matching performance. Unlike traditional methods that rely solely on identity supervision, we draw inspiration from the success of VLP models and leverage the generated textual shape representations as supervisory signals. The signals guide the visual encoder in learning more robust and discriminative body shape features. By explicitly modeling modality-shared shape information in this way, we not only strengthen the model's understanding of high-level semantic features but also effectively reduce modality gaps, thereby improving the robustness and accuracy of cross-modal person re-identification.

Formally, for the visible-infrared image pair ($I_{i}^{v}$, $I_{i}^{r}$), we use the generated identity-level textual shape descriptions as supervisory signals to optimize the backbone network ($\bm E_{v}$, $\bm E_{r}$, $\bm E$) and shape encoder $\bm E_{s}$. Additionally, we leverage the generated textual appearance descriptions to supervise the optimization of the backbone network, promoting the exploration of shared appearance features across modalities. Together, these form a multi-text supervision mechanism that steers the model’s optimization, enabling it to accurately capture identity-related key information across different modalities. This further reduces modality discrepancies and enhances feature discriminability. The corresponding optimization loss is defined as follows:
\begin{equation}
	\begin{aligned}
\mathcal{L}_{i2tce}^{a} =  &- \frac{1}{{{n_v}}}\sum\limits_{i = 1}^{{n_v}} {{\bm y_i}log \frac{{exp({sim( {\bm v^a_{v,i}, \bm t^a_{y_i}})})}}{{\sum\nolimits_{{y_j} = 1}^{N_{y}} {exp({sim({\bm v^a_{v,i}, \bm t^a_{y_j}})})} }}}\\
&- \frac{1}{{{n_r}}}\sum\limits_{i = 1}^{{n_r}} {{\bm y_i}log \frac{{exp({sim( {\bm v^a_{r,i}, \bm t^a_{y_i}})})}}{{\sum\nolimits_{{y_j} = 1}^{N_{y}} {exp({sim({\bm v^a_{r,i}, \bm t^a_{y_j}})})} }}}
	\end{aligned},
 \label{Eq:6}
\end{equation}
\begin{equation}
	\begin{aligned}
\mathcal{L}_{i2tce}^{s} =  &- \frac{1}{{{n_v}}}\sum\limits_{i = 1}^{{n_v}} {{\bm y_i}log \frac{{exp({sim( {\bm v^s_{v,i}, \bm t^s_{y_i}})})}}{{\sum\nolimits_{{y_j} = 1}^{N_{y}} {exp({sim({\bm v^s_{v,i}, \bm t^s_{y_j}})})} }}}\\
&- \frac{1}{{{n_r}}}\sum\limits_{i = 1}^{{n_r}} {{\bm y_i}log \frac{{exp({sim( {\bm v^s_{r,i}, \bm t^s_{y_i}})})}}{{\sum\nolimits_{{y_j} = 1}^{N_{y}} {exp({sim({\bm v^s_{r,i}, \bm t^s_{y_j}})})} }}}
	\end{aligned},
 \label{Eq:7}
\end{equation}
where $N_{y}$ is the total number of identities. By leveraging this multi-text supervision strategy, we not only utilize the prior knowledge of CLIP~\cite{clip}, but also enhance the network's ability to understand and model modality-invariant shape and appearance information. This explicit modeling approach effectively directs the network to focus on identity-related, modality-invariant features, thereby improving the performance of VIReID.

\textbf{Distribution Consistency Constraint}. The losses outlined above guide the network to capture modality-shared identity information by aligning the visual features with their corresponding textual prototypes. To further reduce the distribution gap between cross-modal features, we use identity-level textual representations as reference centers for modality alignment and construct a text-based distribution consistency constraint across modalities. Specifically, for the visible-infrared visual features ($v_{i}^{v}$, $v_{i}^{r}$) and textual prototype set $T_p=\{t_j\}_{j=1}^{N_y}\in \mathbb{R}^{N_y\times d}$, we compute the similarity probability distribution $p_{i,j}$ of $\bm v_{i}$ and $\bm t_{j}$ as follows:
\begin{equation}\
\begin{aligned}\label{Eq:14}
p_{i,j}=\frac{exp(sim(\bm v_i, \bm t_j))}{\sum_{k=1}^{N_y} exp(sim(\bm v_i, \bm t_k))}
\end{aligned},
\end{equation}
where $\bm v_{i} = \{v_{i}^{v}, v_{i}^{r}\}$. When $\bm v_{i}$ is set to $v_{i}^{v}$ and $v_{i}^{r}$ respectively, we obtain $p^v_{i,j}$ and $p^r_{i,j}$. We expect the similarity probability distribution between image features from the visible and infrared modalities and textual prototypes to remain consistent, thereby enhancing cross-modal alignment. This constraint is formulated as follows:
\begin{equation}\
\begin{aligned}\label{Eq:5}
\mathcal{L}_{dcc}=&\frac{1}{n}\sum_{i=1}^{n}\sum_{j=1}^{N_y}(p^v_{i,j}-p^r_{i,j})^2
\end{aligned},
\end{equation}
where $n=n_{v}=n_{r}$. We apply the above constraint to both the shape textual representation $\bm t^s_{j}$ and the appearance textual representation $\bm t^a_{j}$, denoted as $\mathcal{L}^s_{dcc}$ and $\mathcal{L}^a_{dcc}$, respectively. Due to the modality-invariant nature and strong semantic abstraction capability of textual representations, this “distribution-consistent alignment” strategy guides the network to learn modality-shared identity representations without disrupting the intrinsic feature structures of each modality. By enforcing such geometric consistency constraints, we achieve more robust cross-modal identity alignment in the latent semantic space, improving both matching accuracy and the model’s generalization ability.

\subsection{Training and Inference}
The optimization of BSaTa is divided into two stages to ensure the reliability of shape and appearance features while enhancing cross-modal matching performance. In the first stage, we focus on optimizing the learnable textual shape and appearance descriptions $T^s_i$ and $T^a_i$, while freezing the backbone network and shape encoder. The loss function $\mathcal{L}^{s1}=\mathcal{L}^s_{bcon}+\mathcal{L}^a_{bcon}+\lambda_1(\mathcal{L}^s_{tvcr}+\mathcal{L}^a_{tvcr}$) is employed to refine the learnable word tokens, enabling them to accurately capture modality-shared shape and appearance information, thereby generating robust and semantically rich textual representations.

In the second stage, the generated shape and appearance descriptions are frozen, while the backbone network ($\bm E_{v}$, $\bm E_{r}$, $\bm E$) and shape encoder $\bm E_{s}$ are unfrozen for further optimization. In addition to using textual shape and appearance supervision to refine the visual encoder, we also follow~\cite{csdn} and incorporate identity information as an additional supervision signal to enhance the modeling of identity-related features. This includes the identity loss ($\mathcal{L}_{id}$) and the weighted regularized triplet loss ($\mathcal{L}_{wrt}$)~\cite{AGW}. Thus, the overall objective function for the second stage is formulated as:
\begin{equation}\
\begin{aligned}
\mathcal{L}^{s2}=\mathcal{L}_{id}+\lambda_2\mathcal{L}_{wrt}+\lambda_3(\mathcal{L}_{i2tce}^{a}+\mathcal{L}_{i2tce}^{s})+\lambda_4(\mathcal{L}_{dcc}^{s}+\mathcal{L}_{dcc}^{a})
\end{aligned},
\end{equation}
where $\lambda_1$, $\lambda_2$, $\lambda_3$, and $\lambda_4$ balance the contribution of different losses.

The learnable textual descriptions, CLIP’s text encoder, and the semantic parsing model are used exclusively during training to obtain reliable textual descriptions that provide supervision for the visual encoder. These components are removed during inference. During inference, the corresponding appearance and body shape features are extracted from both infrared and visible images. These features are then concatenated, and their cosine similarity score is computed as the retrieval criterion.

\begin{table*}[!t]\footnotesize
\centering {\caption{Performance comparison with state-of-the-art methods on SYSU-MM01. '-' denotes that no reported result is available.}\label{Tab:1}
\renewcommand\arraystretch{1.2}
\begin{tabular}{m{2.0cm}<{\centering}|m{1.5cm}<{\centering}|m{0.6cm}<{\centering}m{0.6cm}<{\centering}|m{0.6cm}<{\centering}m{0.6cm}<{\centering}|m{0.6cm}<{\centering}m{0.6cm}<{\centering}|m{0.6cm}<{\centering}m{0.6cm}<{\centering}}
\hline
 \hline
  \multirow{3}*{Methods} & \multirow{3}*{Venue} & \multicolumn{4}{c|}{All-Search} & \multicolumn{4}{c}{Indoor-Search}\\
  
  \cline{3-10} & & \multicolumn{2}{c|}{Single-Shot} & \multicolumn{2}{c|}{Multi-Shot} & \multicolumn{2}{c|}{Single-Shot} & \multicolumn{2}{c}{Multi-Shot}\\
  
  & & R@1 & mAP & R@1 & mAP & R@1 & mAP & R@1 & mAP\\
\hline



  Hi-CMD \cite{HiCMD} & CVPR'20 & 34.9 & 35.9 & - & - & - & - & - & -\\

  JSIA \cite{JSIA} & AAAI'20 & 38.1 & 36.9
                             & 45.1 & 29.5 
                             & 43.8 & 52.9 
                             & 52.7 & 42.7 \\

  X-Modality \cite{X-modality} & AAAI'20 & 49.9  & 50.7 
                                         & - & - 
                                         & - & - 
                                         & - & -\\


  MSA \cite{msa} & IJCAI'21 & 63.1 & 59.2 
                            & - & - 
                            & 67.1 & 72.7 
                            & - & - \\

  CECNet \cite{CECNet} & TCSVT'22 & 53.3 & 51.8
                                  & - & - 
                                  & 60.6 & 62.8 
                                  & - & - \\

  RBDF \cite{RBDF} & TCYB'22 & 57.6 & 54.4 
                             & - & - 
                             & - & - 
                             & - & - \\

  TSME \cite{TSME} & TCSVT'22 & 64.2 & 61.2 
                              & 70.3 & 54.3 
                              & 64.8 & 71.3 
                              & 76.8 & 65.0 \\

  ACD \cite{acd} & TIFS'24 & 74.4 & 71.1 
                          & 80.4 & 66.9 
                          & 78.9 & 82.7 
                          & 86.0 & 78.6 \\

  AGPI$^2$ \cite{agpi2} & TIFS’25 & 72.2 & 70.5 
                          & - & - 
                          & 83.4 & 84.2 
                          & - & - \\
  \hline







  AGW \cite{AGW} & TPAMI'21 & 47.5 & 47.6 
  & - & - 
  & 54.1 & 62.9 
  & -  & - \\




  MSO \cite{MSO} & MM'21 & 58.7 & 56.4 
  & 65.8 & 49.5 
  & 63.0 & 70.3 
  & 72.0 & 61.6 \\


  
  
  MPANet \cite{MPANet} & CVPR'21 & 70.5 & 68.2 
  & 75.5 & 62.9 
  & 76.7 & 80.9 
  & 84.2 & 75.1 \\

  MMN \cite{mmn} & MM’21 & 70.6 & 66.9 
  & - & - 
  & 76.2 & 79.6 
  & - & - \\
  
  


  FMCNet \cite{FMCNet} & CVPR'22 & 66.3 & 62.5 
  & 73.4 & 56.0 
  & 68.1 & 74.0 
  & 78.8 & 63.8 \\

  DART \cite{dart} & CVPR’22 & 68.7 & 66.2 
  & - & - 
  & 72.5 & 78.1 
  & - & - \\

  DCLNet \cite{DCLNet} & MM'22 & 70.8 & 65.3 
  & - & - 
  & 73.5 & 76.8 
  & - & - \\

  MAUM \cite{MAUM} & CVPR'22 & 71.6 & 68.7 
  & - & - 
  & 76.9 & 81.9 
  & - & - \\





  PMT \cite{PMT} & AAAI'23 & 67.5 & 64.9 
  & - & - 
  & 71.6 & 76.5 
  & - & - \\


  ProtoHPE \cite{protohpe} & MM’23 & 71.9 & 70.5 
  & - & - 
  & 77.8 & 81.3 
  & - & - \\

  CAL \cite{cal} & ICCV’23 & 74.6 & 71.7 
  & 77.0 & 64.8 
  & 79.6 & 83.6 
  & 86.9 & 78.5 \\

  MBCE \cite{mbce} & AAAI'23 & 74.7 & 72.0 
  & 78.3 & 65.7 
  & 83.4 & 86.0 
  & 88.4 & 80.6 \\


  SEFL \cite{SEFL} & CVPR'23 & 75.1 & 70.1 
  & - & - 
  & 78.4 & 81.2 
  & - & - \\


  PMWGCN \cite{PMWGCN} & TIFS'24 & 66.8 & 64.8 
  & - & - 
  & 72.6 & 76.1 
  & - & - \\

  DARD \cite{dard} & TIFS'24 & 69.3 & 65.7 
  & - & - 
  & 79.6 & 60.3 
  & - & - \\


  CAJ$_{+}$ \cite{CAJ+} & TPAMI'24 & 71.4 & 68.1 
  & - & - 
  & 78.4 & 82.0 
  & - & - \\


  RLE \cite{rle} & NeurIPS’24 & 75.4 & 72.4 
  & - & - 
  & 84.7 & \underline{87.0} 
  & - & - \\

  HOS-Net \cite{hosnet} & AAAI’24 & 75.6 & \underline{74.2} 
  & - & - 
  & 84.2 & 86.7 
  & - & - \\

  AMINet \cite{AMINet} & arXiv’25 & 74.7 & 66.1
  & - & - 
  & 79.1 & 82.3 
  & - & - \\
  
  CSCL \cite{cscl} & TMM’25 & 75.7 & 72.0
  & - & - 
  & 80.8 & 83.5 
  & - & - \\

  ScRL \cite{ScRL} & PR’25 & 76.1 & 72.6 
  & - & - 
  & 82.4 & 85.4 
  & - & - \\

  DMPF \cite{dmpf} & TNNLS’25 & 76.4 & 71.5 
  & - & - 
  & 82.2 & 84.9 
  & - & - \\

  CycleTrans \cite{cycletrans} & TNNLS’25 & 76.5 & 72.6 
  & 82.8 & \underline{68.5} 
  & \textbf{87.2} & 84.9 
  & \underline{91.2} & 81.4 \\

  CSDN \cite{csdn} & TMM'25 & \underline{76.7} & 73.0 
  & \textbf{83.5} & 67.9 
  & 84.5 & 86.8 
  & \textbf{91.3} & 82.2 \\


  \hline

  BSaTa (Ours) & - & \bf{77.4} & \textbf{74.3} 
  & \underline{83.2} & \bf{69.6} 
  & \underline{84.7}& \bf{87.1} 
  & 91.0 & \bf{83.2} \\

  \hline\hline
\end{tabular}}
\vspace{-0.3cm}
\end{table*}

\section{Experiments}
\subsection{Experiment Settings}
\textbf{Datasets and Metrics:}
To evaluate the effectiveness of our approach, we conduct experiments on two VIReID datasets. 
\textbf{SYSU-MM01}~\cite{VIREID} is a challenging dataset specifically designed for VIReID, containing 491 identities, with a total of 287,628 visible images and 15,792 infrared images captured using 4 visible-light cameras and 2 infrared cameras in both indoor and outdoor environments. Following the standard evaluation protocol~\cite{VIREID}, the training set consists of 395 identities with 22,258 visible and 11,909 infrared images, while the test set includes 96 identities. The dataset supports two retrieval settings: all-search mode, which includes visible images from both indoor and outdoor cameras, and indoor-search mode, where only indoor camera images are considered. Additionally, testing can be conducted under single-shot and multi-shot gallery settings, where each identity is represented by either 1 or 10 visible images, respectively. \textbf{RegDB}~\cite{RegDB} is a smaller dataset consisting of 8,240 images from 412 identities, with each identity having 10 visible and 10 infrared images captured from a single paired-camera setup. Following the protocol in~\cite{AlignGAN}, 206 identities are randomly selected for training, while the remaining 206 identities form the test set. The dataset allows evaluation in two cross-modal retrieval directions: visible-to-infrared and infrared-to-visible person retrieval. To ensure a fair comparison, each experiment is repeated across 10 random training/testing splits, and the final results are reported as the average over all runs.

For performance evaluation, we adopt standard retrieval metrics, including the Cumulative Matching Characteristics (CMC) curve and mean Average Precision (mAP). Specifically, we report R@1, R@10, and R@20 accuracy to measure retrieval effectiveness at different ranks.

\textbf{Implementation Details:}
For input images, we uniformly resize them to 288$\times$144 and apply standard data augmentation techniques, including random flipping, padding, and cropping, to enhance generalization. Our model adopts CLIP~\cite{clip} as the backbone, incorporating two parallel shallow branches for modality-specific feature extraction, while the remaining four deep convolutional blocks are shared to capture modality-invariant representations. The number of learnable word tokens in the shape and appearance descriptions is set to 4 for both $M$ and $N$. The hyperparameters for different loss components are set as $\lambda_1=0.04$, $\lambda_2=0.15$, $\lambda_3=0.05$ and $\lambda_4=0.05$ to effectively balance the training objectives. Model training utilizes the Adam optimizer with a weight decay factor of 4e-5. The training process is divided into two stages, each with distinct optimization strategies. In the first stage, we train two learnable texts for 60 epochs with an initial learning rate of $3\times10^{-4}$, following a cosine annealing schedule~\cite{clipreid} for gradual decay. In the second stage, we fine-tune the backbone network and shape encoder for 120 epochs. The learning rate is linearly warmed up from $3\times10^{-6}$ to $3\times10^{-4}$ over the first 10 epochs and then decayed by a factor of 0.1 at epochs 40 and 70. The training batch size is set to 64, consisting of 8 identities, with each identity represented by 4 visible and 4 infrared images. We implement our method using PyTorch library and conduct all experiments on a single NVIDIA GTX3090 24GB GPU. 

\vspace{-4pt}
\subsection{Comparisons with State-of-the-art Models}
In this section, we compare our proposed BSaTa framework with the current state-of-the-art (SOTA) methods on two VIReID datasets, as shown in Tables~\ref{Tab:1} and~\ref{Tab:2}. The compared methods are categorized into two groups: generative and non-generative approaches. As observed from the tables, non-generative methods significantly outperform generative ones and currently dominate the VIReID task. Our BSaTa belongs to the non-generative category.

Table \ref{Tab:1} presents the performance comparison with SOTA methods on \textbf{SYSU-MM01}. The proposed BSaTa framework achieves competitive results across all evaluation metrics and settings. Notably, in the commonly used AS–SS (All-Search, Single-Shot) mode, our method achieves SOTA performance, with an impressive R@1 accuracy of 77.4\% and mAP of 74.3\%. Compared to the runner-up method CSDN~\cite{csdn}, BSaTa improves by 0.7\% in R@1 and 1.3\% in mAP. The key difference between BSaTa and CSDN lies in the explicit incorporation of body shape information, and the significant performance gain highlights the effectiveness of our method and the importance of shape information in VIReID. Our approach also performs well in other settings, achieving the highest mAP under AS–MS (All-Search, Multi-Shot), IS–SS (Indoor-Search, Single-Shot), and IS–MS (Indoor-Search, Multi-Shot) modes with scores of 69.6\%, 87.1\%, and 83.2\%, respectively, further demonstrating the robustness and superiority of BSaTa. Table \ref{Tab:2} presents the comparison results on \textbf{RegDB}. Our BSaTa establishes new SOTA performance for visible-to-infrared retrieval, with an R@1 of 95.8\% and mAP of 89.1\%. For infrared-to-visible retrieval, our method achieves competitive performance, being only 0.1\% lower in R@1 compared to MBCE~\cite{mbce}. In summary, BSaTa consistently achieves superior performance across both datasets. This success can be attributed to the effective use of body shape information, which enables the model to extract highly discriminative and modality-invariant identity features.

\begin{table}[!t]\footnotesize
\setlength{\abovecaptionskip}{0.1cm} 
\setlength{\belowcaptionskip}{-0.2cm}
\centering {\caption{Performance comparison with state-of-the-art methods on RegDB.}\label{Tab:2}
\renewcommand\arraystretch{1.2}
\begin{tabular}{m{1.8cm}<{\centering}|m{1.0cm}<{\centering}|cc|cc}
\hline
 \hline
  \multirow{2}*{Methods} & \multirow{2}*{Venue} & \multicolumn{2}{c|}{Visible to Infrared} & \multicolumn{2}{c}{Infrared to Visible} \\
  \cline{3-6} &  & R@1 & mAP & R@1 & mAP\\
  \hline



  JSIA \cite{JSIA} & AAAI’20 & 48.5 & 49.3 & 48.1 & 48.9 \\

  Hi-CMD \cite{HiCMD} & CVPR’20 & 70.9 & 66.0 & - & - \\

  X-Modality \cite{X-modality} & AAAI’20 & - & - & 62.2 & 60.1 \\

  MSA \cite{msa} & IJCAI'21 & 84.8 & 82.1 & - & - \\

  RBDF \cite{RBDF} & TCYB’22 & 79.8 & 76.7 & 76.2 & 73.9 \\

  CECNet \cite{CECNet} & TCSVT’22 & 82.3 & 78.4 & 78.9 & 75.5 \\

  TSME \cite{TSME} & TCSVT’22 & 87.3 & 76.9 & 86.4 & 75.7 \\

  ACD \cite{acd} & TIFS'24 & 84.7 & 83.2 & 87.1 & 84.7 \\

  AGPI$^2$ \cite{agpi2} & TIFS’25 & 89.0 & 83.8 & 87.9 & 83.0 \\

  \hline






   AGW \cite{AGW} & TPAMI’21 & 70.0 & 66.3 & 70.4 & 65.9 \\




  MSO \cite{MSO} & MM’21 & 73.6 & 66.9 & 74.6 & 67.5 \\


  MPANet \cite{MPANet} & CVPR’21 & 83.7 & 80.9 & 82.8 & 80.7 \\

  MMN \cite{mmn} & MM’21 & 91.6 & 84.1 & 87.5 & 80.5 \\

  DCLNet \cite{DCLNet} & MM’22 & 81.2 & 74.3 & 78.0 & 70.6 \\

  DART \cite{dart} & CVPR’22 & 83.6 & 60.6 & 81.9 & 56.7  \\

  MAUM \cite{MAUM} & CVPR'22 & 87.8 & 85.0 & 86.9 & 84.3 \\

  FMCNet \cite{FMCNet} & CVPR'22 & 89.1 & 84.4 & 88.3 & 83.8 \\







  PMT \cite{PMT} & AAAI’23 & 84.8 & 76.5 & 84.1 & 75.1 \\


  ProtoHPE \cite{protohpe} & MM’23 & 88.7 & 83.7 & 88.6 & 81.9 \\

  SEFL \cite{SEFL} & CVPR'23 & 91.0 & 85.2 & 92.1 & 86.5 \\

  MBCE \cite{mbce} & AAAI'23 & 93.1 & 88.3 & \textbf{93.4} & \underline{87.9}  \\





  CAJ$_{+}$ \cite{CAJ+} & TPAMI’24 & 85.6 & 79.7 & 84.8 & 78.5 \\

  DARD \cite{dard} & TIFS'24 & 86.2 & 85.4 & 85.5 & 85.1 \\

  PMWGCN \cite{PMWGCN} & TIFS’24 & 90.6 & 84.5 & 88.7 & 81.6 \\

  RLE \cite{rle} & NeurIPS’24 & 92.8 & \underline{88.6} & 91.0 & 86.6 \\






  DMPF \cite{dmpf} & TNNLS’25 & 88.8 & 81.0 & 88.8 & 81.8 \\

  CycleTrans \cite{cycletrans} & TNNLS’25 & 90.6 & 85.6 & 81.8 & 87.0 \\

  AMINet \cite{AMINet} & arXiv’25 & 91.2 & 84.6 & 89.5 & 82.4 \\
  
  CSCL \cite{cscl} & TMM’25 & 92.1 & 84.2 & 89.6 & 85.0 \\

  ScRL \cite{ScRL} & PR’25 & 92.4 & 86.7 & 91.8 & 85.3 \\

  CSDN \cite{csdn} & TMM'25 & \underline{95.4} & 87.7 & 92.3 & 85.5 \\


  \hline

  BSaTa (Ours) & - & \bf{95.8} & \bf{89.1} & \underline{93.3}  & \bf{87.9} \\

  \hline\hline
\end{tabular}}
\end{table}

\vspace{-4pt}
\subsection{Ablation Studies}
We conduct ablation experiments for BSaTa on SYSU-MM01 under the AS–SS setting.

\textbf{Contributions of Proposed Components:}
Table \ref{Tab:3} presents a comprehensive ablation study evaluating the effectiveness of each component in our BSaTa, including Appearance-Textual Alignment (ATA), Appearance-Aware Representation Learning (AARL), Body Shape-Textual Alignment (BSTA), Text-Visual Consistency Regularization (TVCR), Shape-Aware Representation Learning (SARL), and Distribution Consistency Constraint (DCC).

The baseline model (0\#) refers to the backbone network trained solely with identity loss ($\mathcal{L}_{id}$) and the weighted regularized triplet loss ($\mathcal{L}_{wrt}$), achieving an R@1 accuracy of 68.2\% and a mAP of 64.9\%. When textual appearance descriptions are introduced in Baseline (1\#), leveraging CLIP to enhance the backbone’s ability to model high-level semantic appearance features, the performance experiences a significant boost, with R@1 and mAP increasing by 6.8\% and 7.0\%, respectively. Replacing textual appearance with textual shape descriptions (2\#) also yields a considerable improvement, achieving 75.2\% R@1 and 71.8\% mAP, highlighting the crucial role of shape information in VIReID. When both textual appearance and shape descriptions are jointly utilized (3\#), a multi-text supervision mechanism is formed, leading to further performance enhancement, with gains of 1.0\% in R@1 and 1.6\% in mAP. Building upon this, the addition of the TVCR module (4\#) effectively strengthens visual-textual semantic alignment and ensures the reliability of the generated text embeddings, thereby contributing to improved discriminative capability. Similarly, incorporating the DCC module (5\#) on top of 3\# enables alignment of feature distributions between infrared and visible modalities using text embeddings as semantic anchors. This facilitates the extraction of modality-invariant identity features, resulting in more robust performance. Ultimately, when all modules are jointly applied (6\#), our BSaTa framework achieves the best overall performance across all metrics, demonstrating the synergy between components and the effectiveness of our complete design in addressing the VIReID task.

\begin{table}[!t]\footnotesize
\setlength{\abovecaptionskip}{0.1cm} 
\setlength{\belowcaptionskip}{-0.2cm}
\centering {\caption{Effectiveness of different components on SYSU-MM01.}\label{Tab:3}
\begin{tabular}{c|cccccc|cc}
\hline
\hline
No. & ATA & AARL & BSTA & TVCR & SARL & DCC & R@1 & mAP\\
\hline
0\# &  &   &  &   &  &  & 68.2 & 64.9 \\
1\# &\checkmark  &\checkmark   &  &   &  &  &  75.0 & 71.9 \\
2\# &  &   &\checkmark  &  &\checkmark  &  & 75.2 & 71.8 \\
3\# &\checkmark  &\checkmark   &\checkmark  &  &\checkmark  &  & 76.2 & 73.4 \\
4\# &\checkmark  &\checkmark   &\checkmark &\checkmark  &\checkmark &  & 76.8 & 74.1 \\
5\# &\checkmark  &\checkmark   &\checkmark &  &\checkmark & \checkmark &  76.9 & 73.8 \\
6\# &\checkmark  &\checkmark   &\checkmark &\checkmark  &\checkmark & \checkmark & 77.4 & 74.3 \\
\hline\hline
\end{tabular}}
\vspace{-0.5cm}
\end{table}


\begin{table}[!ht]\small
\setlength{\abovecaptionskip}{0.1cm} 
\setlength{\belowcaptionskip}{-0.2cm}
\centering {\caption{Effects of learnable word token lengths $M$ and $N$ on SYSU-MM01.}\label{Tab:4}
\renewcommand\arraystretch{0.96}
\begin{tabular}{c|cccc}
 \hline\hline
  Method  & R@1 & R@10 & R@20 & mAP \\
  \hline
  $M=2$   & 76.4 & 97.5 & 99.5 &  73.4 \\
  $M=4$ (Ours)   & 77.4 & 97.5 & 99.4 & 74.3 \\
  $M=6$   & 76.4 & 97.7 & 99.5 & 73.4  \\
  $M=8$   & 76.0 & 97.4 & 99.3 &  73.5 \\
  \hline
  $N=2$   & 76.4 & 97.5 & 99.3 &  73.4 \\
  $N=4$ (Ours)   & 77.4 & 97.5 & 99.4 & 74.3 \\
  $N=6$   & 76.5 & 97.6 & 99.4 &  73.6 \\
  $N=8$   & 76.9 & 97.2 & 99.1 &  73.3 \\
  \hline\hline
\end{tabular}}
\vspace{-0.3cm}
\end{table}

\textbf{Impact of learnable word token lengths $M$ and $N$:}
The lengths of learnable word tokens, $M$ and $N$, for textual shape and appearance descriptions are critical parameters that influence the quality of the textual representations. To investigate their impact, we vary $M$ and $N$ in the range of {2, 4, 6, 8} and report the results in Table~\ref{Tab:4}. The observed trend indicates that larger values of $M$ and $N$ generally lead to better performance, as they provide richer contextual information for capturing appearance and shape features. Notably, the model achieves peak performance with an R@1 accuracy of 77.4\% when both $M$ and $N$ are set to 4. However, excessively large values may introduce redundant information, leading to overfitting and increased computational cost. Therefore, we adopt $M = N = 4$ as a trade-off between performance and efficiency.

\begin{table}[!ht]\small
\setlength{\abovecaptionskip}{0.1cm} 
\setlength{\belowcaptionskip}{-0.2cm}
\centering {\caption{Effect of TVCR and DCC on SYSU-MM01.}\label{Tab:5}
\renewcommand\arraystretch{0.96}
\begin{tabular}{c|cccc}
 \hline\hline
  Method  & R@1 & R@10 & R@20  & mAP \\
  \hline
  Base (3\#)   & 76.2 & 96.5 & 98.3 & 73.4  \\
  \hline
  TVCR-A   & 76.4 & 96.9 & 98.7 & 73.7 \\
  TVCR-S   & 76.5 & 97.1 & 98.5 & 73.6 \\
  TVCR-SA (Ours) & 76.8 & 97.5 & 99.3 & 74.1 \\
  \hline
  DCC-A    & 76.7 & 97.2 & 99.1 & 74.0 \\
  DCC-S    & 76.4 & 96.8 & 98.9 & 73.8 \\
  DCC-SA (Ours) & 76.9 & 97.3 & 99.3 & 73.8 \\
  \hline\hline
\end{tabular}}
\vspace{-0.3cm}
\end{table}

\textbf{Effect of TVCR and DCC:}
In this section, we evaluate the effectiveness of the Text-Visual Consistency Regularizer (TVCR) and the Distribution Consistency Constraint (DCC), as shown in Table~\ref{Tab:5}. Specifically, based on model 3\# in Table~\ref{Tab:3}, we examine the performance impact of applying these modules individually to either the appearance or shape textual representations, as well as simultaneously to both. The results show that applying TVCR or DCC to a single textual representation yields distinct performance gains, confirming the effectiveness of our proposed modules. Furthermore, the best performance is achieved when both modules are applied together to both textual shape and appearance representations.

\section{Conclusion}
In this paper, we proposed an effective Body Shape-aware Textual Alignment (BSaTa) framework to address the modality gap and semantic limitations in VIReID. Unlike prior works that overlook the importance of body shape information, our approach explicitly models and leverages body shape cues through structured textual representations. The proposed BSTA module effectively extracts body shape semantics using human parsing and CLIP-based encoding, while the TVCR ensures alignment between textual and visual body shape features. Furthermore, the integration of Shape-aware Representation Learning (SRL), incorporating multi-text supervision and distribution consistency constraint, enables the visual encoder to learn more discriminative and modality-invariant identity representations. Extensive experiments on the SYSU-MM01 and RegDB datasets demonstrate that our method achieves superior performance, validating the effectiveness of body shape modeling and our BSaTa. In the future, we plan to explore more fine-grained and dynamic textual guidance to further enhance VIReID.


\clearpage
\bibliographystyle{ACM-Reference-Format}
\bibliography{sample-base}


\end{document}


\title{Prototypical Prompting for Text-to-image Person Re-identification}


\author{Shuanglin Yan}
\affiliation{%
  \institution{Nanjing University of Science and Technology}
  \state{Nanjing}
  \country{China}
}
\email{shuanglinyan@njust.edu.cn}

\author{Jun Liu}
\affiliation{%
  \institution{Singapore University of Technology and Design}
  \streetaddress{8 Somapah Rd}
  \country{Singapore}}
\email{jun_liu@sutd.edu.sg}

\author{Neng Dong}
\affiliation{%
  \institution{Nanjing University of Science and Technology}
  \state{Nanjing}
  \country{China}}
\email{neng.dong@njust.edu.cn}

\author{Liyan Zhang}
\affiliation{%
  \institution{Nanjing University of Aeronautics and Astronautics}
  \state{Nanjing}
  \country{China}}
\email{zhangliyan@nuaa.edu.cn}
  
\author{Jinhui Tang}
\affiliation{%
  \institution{Nanjing University of Science and Technology}
  \state{Nanjing}
  \country{China}}
\email{jinhuitang@njust.edu.cn}

\renewcommand{\shortauthors}{Shuanglin Yan, et al.}

\begin{abstract}
In this paper, we study the problem of Text-to-Image Person Re-identification (TIReID), which aims to find images of the same identity described by a text sentence from a pool of candidate images. Benefiting from Vision-Language Pre-training, such as CLIP (Contrastive Language-Image Pretraining), the TIReID techniques have achieved remarkable progress recently. However, most existing methods only focus on instance-level matching and ignore identity-level matching, that is, many-to-many matching between images and texts under the same identity. In this paper, we propose a novel prototypical prompting framework (Propot) designed to simultaneously model instance-level and identity-level matching for TIReID. Our Propot transforms the identity-level matching problem into a prototype learning problem, which aims to learn identity-enriched prototypes. Specifically, Propot works by ‘initialize, adapt, enrich, then aggregate’. We first utilize CLIP to produce initial prototypes to ensure their quality. Then, we propose a domain-conditional prototypical prompting (DPP) module to prompt the initial prototypes for adapting to the TIReID task through task-related information. Further, we propose an instance-conditional prototypical prompting (IPP) module to update prototypes conditioned on intra-modal and inter-modal instances to ensure prototype diversity. Finally, we design an adaptive prototype aggregation module to aggregate the above generated prototypes to generate the final identity-enriched prototypes. With identity-enriched prototypes, we diffuse the rich identity information of prototypes to instances through prototype-to-instance contrastive loss to model identity-level matching. Extensive experiments on three benchmarks demonstrate the superiority of Propot against existing TIReID methods.
\end{abstract}

\ccsdesc[100]{Information systems~Top-k retrieval in databases}

\keywords{Text-to-image person re-identification, Identity-level matching, Prototypical prompting}

\maketitle

\begin{figure}[!t]
\setlength{\abovecaptionskip}{0.1cm}
\setlength{\belowcaptionskip}{-0.4cm}
\subfigbottomskip=-2pt
\subfigcapskip=-2pt
\centering
\subfigure[Example of TIReID data]{\includegraphics[height=1.6in,width=3.2in]{images/motivation-a.pdf}}
\hspace{-10mm}
\subfigure[Existing matching paradigm] {\includegraphics[height=0.6in,width=3.2in]{images/motivation-b.pdf}}
\subfigure[Our proposed Propot] {\includegraphics[height=1.0in,width=3.2in]{images/motivation-c.pdf}}
\caption{The motivation of our proposed Propot. (a) Some examples of TIReID data containing multiple images from two identities and their annotated texts. Instances under the same identity are significantly different. (b) Most existing TIReID methods only focus on instance-level matching and ignore identity-level matching. (c) Our Propot proposes a prototype prompting framework to produce identity-enriched prototypes and diffuse their rich identity information to instances for modeling identity-level matching.}
\label{Fig:1}
\end{figure}

\section{Introduction}
Person re-identification (ReID), devoted to searching a person-of-interest across different times, locations, and camera views, has garnered increasing interest due to its huge practical value. In recent years, we have witnessed great progress on image-to-image person re-identification (IIReID)~\cite{li2023logical, AIESL, pifhsa, cdvr, GLMC}, which has been successfully applied in various practical scenarios. However, ReID is still challenging when pedestrian images from some cameras are missing. In contrast, textual descriptions are more readily accessible and freer than images collected by specialized equipment, which can be obtained from witnesses at the scene. Thus, text-to-image person re-identification (TIReID)~\cite{GNA} has received a lot of research attention recently owing to it being closer to real-world scenarios.

Compared with IIReID, the critical challenge of TIReID is to solve the modal gap between images and texts while modeling their correspondence. Various methods have been devised to address this challenge by cross-modal interactive attention mechanisms~\cite{HGAN, MIA, DSSL}, semantically aligned local feature learning~\cite{SSAN, tipcb, MANet, lgur}, and fine-grained auxiliary task-enhanced global feature learning~\cite{C2A2, ASAMN}. Nowadays, the emergence of vision-language pre-training models (VLP) has propelled the advancements in computer vision, showcasing exceptional capabilities in semantic understanding, multi-modal alignment, and generalization. It is intuitive to consider that the rich multi-modal prior information of VLP can be harnessed to enhance TIReID models in modeling inter-modal correspondence. Some follow-up CFine~\cite{CFine}, IRRA~\cite{irra} extend VLP for TIReID, greatly promoting the progress of this task. 

Despite the success of VLP-based methods, the current performance of TIReID lags behind that of IIReID, especially in complex scenes such as multiple cameras and drastic view variations. This is mainly attributed to the specificity of the TIReID task. As shown in Figure~\ref{Fig:1}(a), there are multiple images of the same identity, and each image is annotated with a specific text at the instance level. The purpose of TIReID is to find all images under the same identity given a text, that is, to correctly match each text with all images under the same identity. However, most existing methods limited by the alignment objective function~\cite{LCR2S} are trained to only consider \textbf{instance-level matching} between each image and its annotated text, ignoring matches with other non-annotated texts of the same identity, called \textbf{identity-level matching}. An intuitive idea to this problem is to directly match all images and texts under the same identity, but this crude strategy can disrupt instance-level matching relationship~\cite{LCR2S} due to huge view variations, leading to performance collapse. To date, there have been limited solutions~\cite{SSAN, LCR2S} proposed to model identity-level matching for TIReID. SSAN~\cite{SSAN} introduces an enhanced alignment loss to simultaneously align the image with its annotated text and another text sharing the same identity. LCR$^2$S~\cite{LCR2S} designs a teacher-student network architecture to randomly fuse two images/texts under the same identity and then align them to model identity-level matching. While effective, they do not comprehensively model the matching relationship between all images and texts under the same identity. Moreover, the two-stage teacher-student framework incurs high training costs and lacks practicality. In this paper, we thus ask: \emph{Can an end-to-end efficient TIReID model be trained to simultaneously model instance-level and identity-level matching?}

To address this problem, we propose a novel \textbf{Pro}totypical \textbf{p}r\textbf{o}mp-\\\textbf{t}ing framework (\textbf{Propot}) that enables the network to model instance-level and identity-level cross-modal matching for TIReID simultaneously. As shown in Figure \ref{Fig:1}, existing TIReID methods only model instance-level matching between each image and its labeled text. In contrast, our Propot framework introduces identity-level matching based on the above instance-level matching. We learn a prototype containing rich identity information for each identity, and diffuse the rich information contained in the prototype to each instance, indirectly modeling identity-level matching. This transforms the identity-level matching problem into an \textbf{identity-enriched prototype learning} problem. 

Specifically, our Propot follows the ‘initialize, adapt, enrich, then aggregate’ pipeline. Given the powerful multi-modal alignment capability of CLIP, we first cluster the instance (image/text) features extracted by CLIP under each identity as the initial prototype of each identity. Due to the domain gap between CLIP and TIReID training data, the initial prototype was not fully adapted to TIReID. To solve this problem, inspired by CoOp~\cite{coop}, we propose a domain-conditional prototypical prompting (DPP) module, which introduces a set of learnable prompt tokens to learn target domain knowledge and pass it to the initial prototype to adapt it to the TIReID task. However, instances under the same identity exhibit significant differences and diversity due to intrinsic factors such as view changes and camera parameters. If this diversity is ignored, the learned prototype will tend to be monotonous, and rich identity information will be lost. Therefore, inspired by CoCoOp~\cite{cocoop}, we propose an instance-condition prototypical prompting (IPP) module, which makes a prototypical prompt conditioned on a batch of instances. To further enrich the diversity of prototypes, we use both intra-modal instances and inter-modal instances to make two prototypical prompts, further narrowing the modal gap. Afterward, to fully integrate the multiple prototypes generated above, an adaptive prototype aggregation (APA) module is designed, which regards the initial prototype generated by CLIP as the prototype baseline and adaptively ensemble these generated prototypes as the final prototype. Finally, we utilize the prototype-to-instance contrastive loss to diffuse the rich identity information of the prototype to each instance, enabling the network to effectively model identity-level matching. Note that our Propot is single-stage and end-to-end trainable. During inference, only the backbone of the network is used for inference, which is simple and efficient. 

The main contributions of this paper are summarized as follows: (1) We propose an end-to-end trainable prototypical prompting framework to model instance-level and identity-level cross-modal matching for TIReID simultaneously. (2) We transform the identity-level matching problem into an identity-enriched prototype learning problem. Benefiting from CLIP and prompt learning, we utilize CLIP to generate initial prototypes and propose a domain-conditional prototypical prompting (DPP) module and an instance-condition prototypical prompting (IPP) module to generate identity-enriched prototypes. And an adaptive prototype aggregation (APA) module is designed to fully integrate the multiple prototypes. (3) Extensive experiments have been conducted to validate the effectiveness of Propot, and it achieves superior performance on the CUHK-PEDES, ICFG-PEDES, and RSTPReid benchmarks.

\vspace{-0.21cm}
\section{Related Work}
\subsection{Text-to-Image Person Re-identification}
TIReID~\cite{GNA} has been of growing interest over the last few years. Existing studies in TIReID can be broadly divided into the following categories: better model architectures and optimization losses, better alignment strategies, and richer prior information. Previous methods~\cite{Dual, CMPM, CMKA} have focused on designing network and optimization loss to learn globally aligned image and text features in a joint embedding space. These methods are simple and efficient, but ignore detailed information and fine-grained correspondences. To address the limitations of preceding approaches, subsequent methods have been dedicated to refining matching strategies to mine fine-grained correspondence between modalities. Some methods~\cite{GNA, HGAN, MIA, PMA, mgel, DSSL} have emerged to perform fine-grained matching through pairwise interactions between local parts of images and texts. While these approaches demonstrate superior performance by fostering ample cross-modal interactions, they come at the expense of escalating computational costs. To mitigate computational overhead, certain subsequent works~\cite{SSAN, tipcb, LapsCore} adopt local image parts as references to guide the generation of locally aligned text features, avoiding the pairwise interactions between them. However, the efficacy of these methods is constrained by the quality of explicitly acquired local parts. Consequently, alternative methods~\cite{MANet, saf, ivt, lgur} propose diverse aggregation schemes to adaptively aggregate images and text into a set of modality-shared local features, avoiding explicit local part acquisition. 


Recently, visual-language pre-training models (VLP)~\cite{clip} have made significant progress. To benefit from its rich multi-modal prior information, some recent state-of-the-art methods~\cite{TextReID, CFine, irra, empirical} have proposed diverse strategies to tailor VLP for TIReID, resulting in notable performance enhancements. Despite the substantial progress in TIReID, existing methods overlook the inherent nature of TIReID as an identity-level multi-view image-text matching challenge, distinct from a conventional instance-level image-text matching problem. Addressing this concern, LCR$^2$S~\cite{LCR2S} devised a teacher-student network to reason about a comprehensive representation containing multi-view information from a single image or text. This approach yielded significantly improved performance, albeit with limitations in fully integrating multi-view information and exhibiting lower training efficiency. In this study, we present an end-to-end framework to learn a prototype with comprehensive information for each identity based on the entire training set, and transfer multi-view information to individual samples for identity-level matching, providing a solution that enhances both efficiency and the integration of multi-view data for TIReID.

\vspace{-4pt}
\subsection{Vision-Language Pre-Training}
Nowadays, the "pre-training and fine-tuning" paradigm stands as a foundational approach in the computer vision community, in which pre-training models that can provide rich prior knowledge for various downstream vision tasks have gained increasing attention. Previous prevailing practice is rooted in the supervised unimodal pre-training~\cite{vit, resnet} on ImageNet~\cite{imagenet}, followed by fine-tuning on various downstream tasks. In response to the need for enhanced representation capabilities and to overcome annotation constraints, a novel paradigm known as language-supervised vision pre-training (vision-language pre-training, VLP) has emerged. Within VLP, investigating the interaction between vision and language has become a central research focus. Several works~\cite{vilbert, visualbert, uniter, albef, beit} have been proposed to model the interaction of vision and language based on some multi-modal reasoning tasks, such as masked language/region modeling, and image captioning. Recently, contrastive representation learning~\cite{clip, align, filip}, which learns representations by contrasting positive pairs against negative pairs, has gotten well studied for diverse visual representation learning. The representative work, Contrastive Language-Image Pretraining (CLIP)~\cite{clip}, has strong multi-modal semantic representation and zero-shot generalization capabilities, which is trained on 400 million image-text pairs. CLIP has shown promising adaptability for various downstream tasks,  such as video-text retrieval~\cite{clip4clip, clip2video}, referring image segmentation~\cite{cris}, and person re-identification~\cite{CFine}. Our work is built upon the CLIP and leverages its ample multi-modal prior information for learning identity-enriched prototypes.

\vspace{-4pt}
\subsection{Prompt Learning}
Prompt learning~\cite{autoprompt, how} originates from the natural language processing (NLP) domain, which aims to adapt pre-training models to various downstream tasks through prompting. Prompts are typically instructions in the form of sentences that are given to the pre-training model to better understand the downstream task. Early prompts were manually designed for a specific task to better understand the task. However, due to issues such as instability and knowledge bias, recent studies have introduced the concept of prompt learning, wherein task-adapted prompts are automatically generated during the fine-tuning stage. Currently, prompt learning has been extended to the computer vision domain. CoOp~\cite{coop} pioneered the application of prompt learning to adapt large vision-language models in computer vision, while CoCoOp~\cite{cocoop} built upon CoOp to introduce a conditional prompt learning framework, addressing challenges related to weak generalization. Chen et al.~\cite{pvu} proposed to efficiently adapt the CLIP model to the resource-hungry video understanding task by optimizing a few continuous prompt vectors. CLIP-ReID~\cite{clipreid} designed a two-stage framework to generate coarse descriptions of pedestrians to fully utilize the powerful capabilities of CLIP for the ReID task. Inspired by these, in this work we propose to leverage prompt learning to learn comprehensive prototype representations to model many-to-many matching for TIReID.

\begin{figure*}[t!]
  \centering
  \setlength{\abovecaptionskip}{0.1cm}
  \setlength{\belowcaptionskip}{-0.2cm}
  \includegraphics[width=\linewidth,height=3.8in]{images/framework.pdf}\\
  \caption{Overview of our Propot. It includes instance-level matching and identity-enriched prototype learning. For instance-level matching, each image and its annotated text are directly aligned through SDM loss (Baseline). For prototype learning, we first utilize pre-trained CLIP to generate the initial prototypes ($\bm {pt}^v$ and $\bm {pt}^t$). We then adapt the initial prototypes to TIReID through the DPP module to generate the task-adapted prototypes ($\bm {p}_a^v$ and $\bm {p}_a^t$). And the IPP module updates the prototypes conditioned on a batch of intra-modal and inter-modal instances to generate intra-modal and inter-modal enriched prototypes ($\bm {p}_{en}^v$, $\bm {p}_{en}^t$, $\bm {p}_{eo}^v$ and $\bm {p}_{eo}^t$), respectively. The above multiple prototypes are aggregated through Adaptive Prototypical Aggregation (APA) to generate the final prototypes ($\bm {p}^v$ and $\bm {p}^t$), and their rich identity information is diffused to each instance through prototype-to-instance contrastive loss ($\mathcal{L}_{p2v}$, $\mathcal{L}_{p2t}$) to model identity-level matching. Moreover, we also introduce the MLM module as compensation to model fine-grained matching. During testing, only visual and textual encoders are used for inference.}
  \label{Fig:2}
\end{figure*}

\section{The Propot Framework}
Our Propot is a conceptually simple end-to-end trainable framework, and the overview is depicted in Figure~\ref{Fig:2}. We first extract features of each image and text instance through visual and textual encoders, followed by instance-level and identity-level matching. For instance-level matching, we follow existing matching paradigms by directly minimizing the distance between each image and its annotated text. For identity-level matching, we transform it into an identity-enriched prototype learning problem. The aim is to learn a rich prototype containing all instance identity information for each identity in the training set. 
Further details on Propot are expounded in the subsequent subsections.

\subsection{Feature Extraction}
Prior studies~\cite{CFine, irra, empirical} have underscored the effectiveness of CLIP~\cite{clip} in addressing the challenges of TIReID. To harness the extensive multi-modal prior knowledge embedded in CLIP, we leverage CLIP’s image and text encoders to initialize Propot’s image and text backbones. Concretely, given an image-text pair $(I, T)$, we exploit CLIP pre-trained ViT model for extracting visual representations for the image $I$, resulting in the global visual feature $\bm v \in \mathbb{R}^{d}$. For the text caption $T$, the CLIP's textual encoder is utilized to generate the global textual feature $\bm t \in \mathbb{R}^{d}$.


The instance-level matching is modeled by directly aligning $\bm v$ and $\bm t$. As previously highlighted, to model identity-level matching, it becomes imperative to generate a prototype containing rich identity information for each identity. In Propot, the prototype learning method plays a pivotal role. While it is conceivable to learn a prototype for each identity from scratch, akin to previous methodologies~\cite{clipreid}, such an approach introduces challenges in network convergence and does not guarantee prototype quality. To address these concerns, we introduce a novel ‘initialize, adapt, enrich, then aggregate’ prototype learning scheme, detailed in the subsequent subsections.

\subsection{Initial Prototype Generation} 
Leveraging the powerful semantic information extraction and multi-modal alignment capabilities inherent in the CLIP model, we initiate our approach by employing CLIP to extract features for each instance within the TIReID training set. Subsequently, we cluster these instance features based on shared identity labels to produce initial prototypes for the respective modalities. Specifically, given a TIReID training set $\{I_i, T_i, Y_i\}_{i=1}^{N_s}$, where $Y_i \in\{L_1, L_2, ..., L_N\}$ represents the identity label of the image-text pair $(I_i, T_i)$, $N_s$ denotes the number of image-text pairs, and $N$ denotes the number of identity label, we employ the pre-trained CLIP visual and textual encoders to obtain the visual and textual features $\{\bm v_{i}, \bm t_{i}\}_{i=1}^{N_s}$ of all image-text pairs in the training set. We then perform feature aggregation on the image and text features. Taking identity $L_i$ as an example, We generate initial prototypes $\bm {pt}_i^v$ and $\bm {pt}_i^t$ for identity $L_i$ as follows:
\begin{equation}\
\begin{aligned}
\bm {pt}_i^v=\sum_{j=1,Y_j \in L_i}^{N_i}\bm v_{j}, \quad \bm {pt}_i^t=\sum_{j=1,Y_j \in L_i}^{N_i}\bm t_{j}
\end{aligned},
\end{equation}

where $N_i$ denotes the number of instances under the identity $L_i$. Therefore, for the entire training set, we can generate visual and textual initial prototype sets $\bm {pt}^v=[\bm {pt}_1^v, \bm {pt}_2^v, ..., \bm {pt}_N^v]\in \mathbb{R}^{N\times d}$ and $\bm {pt}^t=[\bm {pt}_1^t, \bm {pt}_2^t, ..., \bm {pt}_N^t]\in \mathbb{R}^{N\times d}$.

\subsection{Domain-conditional Prototypical Prompting}
Capitalizing on the potent capabilities of CLIP, the initial prototype effectively captures certain identity information to model identity-level matching, leading to improved performance in TIReID (as shown in Table IV), which also confirms the viability of our idea. However, a notable challenge arises due to the domain gap between the pre-training data of CLIP and the TIReID dataset. Consequently, the identity information mined by CLIP's features falls short, significantly diminishing the impact of the initial prototype. In response to this challenge and drawing inspiration from the Contextual Optimization (CoOp) framework~\cite{coop}, which introduces a set of learnable context vectors to adapt VLMs to downstream tasks, we introduce a Domain-conditional Prototypical Prompting (DPP) module to adapt the initial prototype to the TIReID task. Specifically, we incorporate a set of learnable contextual prompt vectors preceding each initial prototype. These vectors undergo training on the TIReID dataset concurrently with the network, acquiring domain knowledge specific to the TIReID task. These vectors then transmit the domain knowledge to each initial prototype through the self-attention encoder (SAE), facilitating the adaptation of prototypes to the TIReID task.

Formally, for each initial prototype $\bm {pt}_i$, we add a set of learnable contextual prompt vectors $\{[\bm X_i]_1, [\bm X_i]_2, ..., [\bm X_i]_K\}\in \mathbb{R}^{K\times d}$ in front of it, where $[\bm X_i]$ is the visual contextual prompt vector $\bm c_{v,i}$ for the visual prototype $\bm {pt}_i^v$, and $[\bm X_i]$ is the textual contextual prompt vector $\bm c_{t,i}$ for the textual prototype $\bm {pt}_i^t$. Then, we feed $\{[\bm X_i]_1, [\bm X_i]_2, ..., [\bm X_i]_K, \bm {pt}_i\}\in \mathbb{R}^{(K+1)\times d}$ into the self-attention encoder (SAE) to pass the information to the initial prototype for updating it.
\begin{equation}\
\begin{aligned}
\bm p_{a,i}=SAE(\{[\bm X_i]_1, [\bm X_i]_2, ..., [\bm X_i]_K, \bm {pt}_i\})
\end{aligned},
\end{equation}
where $\bm p_{a,i}\in \mathbb{R}^{d}$ represents the task-adaptive prototype. SAE is comprised of $N_a$ blocks, with each block containing a multi-head self-attention layer and a feed-forward network layer. The above process allows us to generate modality-specific task-adaptive prototypes, denoted as $\bm p^v_{a,i}$ and $\bm p^t_{a,i}$, respectively, according to the input prototypes of different modalities.

\subsection{Instance-conditional Prototypical Prompting}
So far, we have generated a prototype adapted to the TIReID task. As expected, the resulting prototype resulted in significant performance improvements. However, as depicted in Figure 1, instances under the same identity showcase notable differences and diversity attributable to intrinsic factors like view changes and camera parameters. The previously derived prototype fails to account for this diversity, resulting in the loss of certain discriminative identity information. To solve this problem, inspired by Conditional Context Optimization (CoCoOp)~\cite{cocoop}, we propose an Instance-conditional Prototypical Prompting (IPP) module. This module serves to update the prototype conditioned on each batch of the input instances, allowing for the comprehensive mining of instance-specific information and ensuring diversity within the prototype.

Specifically, for a batch of $B$ image-text pairs $\{I_i, T_i\}_{i=1}^B$, after feature encoding, the encoded features can be expressed as $\bm V_B=[\bm v_{1}, \bm v_{2}, ..., \bm v_{B}]\in \mathbb{R}^{B\times d}$ and $\bm T_B=[\bm t_{1}, \bm t_{2}, ..., \bm t_{B}]\in \mathbb{R}^{B\times d}$. Subsequently, the instance features $\bm V_B/\bm T_B$ and the prototypes $\bm {pt}^v/\bm {pt}^t$ undergo processing in a cross-attention decoder (CAD). This decoder facilitates the interaction between instance features and prototypes, allowing for the extraction of identity-relevant information from instances to enrich the prototypes. There are $N_e$ blocks in CAD, each featuring a multi-head cross-attention layer and a feed-forward network layer. For the cross-attention decoder, the prototypes $\bm {pt}^v/\bm {pt}^t$ serve as \emph{query}, while the instance features $\bm V_B/\bm T_B$ are designated as \emph{key} and \emph{value}, which is formulated as follows:
\begin{equation}\
\begin{aligned}
\bm p_{en,i}^v=CAD(\bm {pt}^v, \bm V_B, \bm V_B),\quad\bm p_{en,i}^t=CAD(\bm {pt}^t, \bm T_B, \bm T_B)
\end{aligned},
\end{equation}
where $\bm p_{en,i}^v\in \mathbb{R}^{d}$ and $\bm p_{en,i}^t\in \mathbb{R}^{d}$ represent the intra-modal instance-enriched visual and textual prototypes, respectively. The above leverages the intra-modal instance features to enrich the prototypes. Additionally, we extend this enrichment by incorporating inter-modal instance features, fostering the network to extract modality-shared identity information comprehensively and minimize the modal gap. We formulate them as
\begin{equation}\
\begin{aligned}
\bm p_{eo,i}^v=CAD(\bm {pt}^v, \bm T_B, \bm T_B),\quad \bm p_{eo,i}^t=CAD(\bm {pt}^t, \bm V_B, \bm V_B)
\end{aligned},
\end{equation}
where $\bm p_{eo,i}^v\in \mathbb{R}^{d}$ and $\bm p_{eo,i}^t\in \mathbb{R}^{d}$ represent the inter-modal instance-enriched visual and textual prototypes, respectively.

\subsection{Adaptive Prototype Aggregation}
Consequently, for both images and texts, we generate three distinct modality-specific prototypes: $\bm p_{a,i}^m\in \mathbb{R}^{d}$, $\bm p_{en,i}^m\in \mathbb{R}^{d}$, and $\bm p_{eo,i}^m\in \mathbb{R}^{d}$, each containing different information for each identity $L_i$, where $m\in \{v, t\}$. To seamlessly integrate all available information, we introduce an Adaptive Prototype Aggregation (APA) module. This module effectively aggregates diverse prototypes to construct a comprehensive identity-enriched prototype. As $\bm {pt}_i^m$ is directly derived by CLIP, benefiting from its strong semantic understanding capabilities, the quality of the prototype is assured. We designate $\bm {pt}i^m$ as the prototype baseline for the aggregation process and the weights for aggregation are determined by the correlation of other prototypes with $\bm {pt}i^m$. This approach enables the suppression of spurious identity prototype information in $\bm p{a,i}^m, \bm p{en,i}^m, \bm p_{eo,i}^m$, while amplifying the correct ones during aggregation. The calculation of prototype correlation, serving as the aggregation weights for the three prototypes, is articulated as follows: 
\begin{equation}\
\begin{aligned}
\bm w_{a,i}^m=\bm {pt}_i^m(\bm p_{a,i}^m)^T,\quad\bm w_{en,i}^m=\bm {pt}_i^m(\bm p_{en,i}^m)^T,\quad\bm w_{en,i}^m=\bm {pt}_i^m(\bm p_{en,i}^m)^T
\end{aligned}.
\end{equation}

Then, we utilize the softmax function to normalize the weights and obtain the final identity-enriched prototype as
\begin{equation}\
\begin{aligned}
\bm p_i^m=\bm {pt}_i^m+\sum_{k}\bm w_{k,i}^m\cdot\bm p_{k,i}^m
\end{aligned},
\end{equation}
where $k\in \{a, en, eo\}$. Thus, for the entire training set, we can generate the final visual and textual prototype sets $\bm {p}^v=[\bm {p}_1^v, \bm {p}_2^v, ..., \bm {p}_N^v]\in \mathbb{R}^{N\times d}$ and $\bm {p}^t=[\bm {p}_1^t, \bm {p}_2^t, ..., \bm {p}_N^t]\in \mathbb{R}^{N\times d}$. Subsequently, we propagate the rich identity information encapsulated in the prototypes to each instance through prototype-to-instance contrastive loss to model identity-level matching for TIReID.

\subsection{Training and Inference}
The goal of Propot is to model both instance-level and identity-level matching for TIReID. To this end, we optimize Propot through cross-modal matching loss, cross-entropy loss, prototype-to-instance contrastive loss, and mask language modeling loss. 

Given a batch of $B$ image-text pairs $\{I_i, T_i\}_{i=1}^B$, we generate the global visual and textual features as $\bm V_B=[\bm v_{1}, \bm v_{2}, ..., \bm v_{B}]\in \mathbb{R}^{B\times d}$ and $\bm T_B=[\bm t_{1}, \bm t_{2}, ..., \bm t_{B}]\in \mathbb{R}^{B\times d}$. To align each image $I_i$ and its annotated text $T_i$, we utilize the similarity distribution matching (SDM)~\cite{irra} as the cross-modal matching loss to model instance-level matching between them.
\begin{equation}\
\begin{aligned}
\mathcal{L}_{sdm}=\mathcal{L}_{i2t}+\mathcal{L}_{t2i}
\end{aligned},
\end{equation}
\begin{equation}\
\begin{aligned}\label{Eq:13}
\mathcal{L}_{i2t}=\frac{1}{B}\sum_{i=1}^{B}\sum_{j=1}^{B}p_{i,j}log(\frac{p_{i,j}}{q_{i,j}+\epsilon})
\end{aligned},
\end{equation}
\begin{equation}\
\begin{aligned}\label{Eq:14}
p_{i,j}=\frac{exp(sim(\bm v_i, \bm t_j)/\tau)}{\sum_{k=1}^Bexp(sim(\bm v_i, \bm t_k)/\tau)}
\end{aligned},
\end{equation}
where $sim(\cdot)$ denotes the cosine similarity function, $\tau$ denotes the temperature factor to control the distribution peaks, and $\epsilon$ is a small number to avoid numerical problems. $q_{i,j}$ denotes the true matching probability. $\alpha$ indicates the margin. $\mathcal{L}_{t2i}$ can be obtained by exchanging $\bm v$ and $\bm t$ in Eqs.~\ref{Eq:13} and \ref{Eq:14}. Moreover, to ensure the discriminability of features $\bm v$ and $\bm t$, we calculate cross-entropy loss $\mathcal{L}_{id}$ on them to classify them into corresponding identity labels.

Through the prototype learning process described above, we generate two identity-rich prototypes  $\bm {p}_i^v$ and $\bm {p}_i^t$ for each identity $L_i$. To effectively diffuse the rich identity information encapsulated in the prototypes to instances of the same identity, we employ a prototype-to-instance contrastive loss, denoted as $\mathcal{L}_{p2i}$. This loss operates in tandem with the cross-modal matching loss, collectively contributing to the modeling of identity-level matching.
\begin{equation}\
\begin{aligned}
\mathcal{L}_{p2i}=\sum_{i=1}^{N}\mathcal{L}_{p2v}(L_i)+\mathcal{L}_{p2t}(L_i)
\end{aligned},
\end{equation}
\begin{equation}\
\begin{aligned}
\mathcal{L}_{p2v}(L_i)=-\frac{1}{|P(L_i)|}\sum_{p\in P(L_i)}\frac{exp(sim(\bm v_p, \bm p_i^v)/\tau}{\sum_{k=1}^Bexp(sim(\bm v_j, \bm p_i^v)/\tau)}
\end{aligned},
\end{equation}
\begin{equation}\
\begin{aligned}
\mathcal{L}_{p2t}(L_i)=-\frac{1}{|P(L_i)|}\sum_{p\in P(L_i)}\frac{exp(sim(\bm t_p, \bm p_i^t)/\tau}{\sum_{k=1}^Bexp(sim(\bm t_j, \bm p_i^t)/\tau)}
\end{aligned},
\end{equation}
where $P(L_i)$ represents a set of instance indices of identity $L_i$, and $|P(L_i)|$ denotes the cardinality of $P(L_i)$. To further improve TIReID performance, we follow~\cite{irra} to introduce a mask language modeling task denoted by $\mathcal{L}_{mlm}$ to model fine-grained matching between modalities.

Propot is a single-stage and end-to-end trainable framework, and the overall objective function $\mathcal{L}$ for training is as follows:
\begin{equation}\
\begin{aligned}
\mathcal{L}=\mathcal{L}_{sdm}+\mathcal{L}_{id}+\lambda_1\mathcal{L}_{p2i}+\lambda_2\mathcal{L}_{mlm}
\end{aligned},
\end{equation}
where $\lambda_1$ and $\lambda_2$ balance the contribution of different loss terms during training.

After training, Propot has the ability of instance-level and identity-level matching. The prototype learning process only exists during the training phase. In the inference phase, we only use Propot’s visual and textual encoders to extract the global features of the test samples. The retrieval results are then obtained by calculating the cosine similarity. 

\section{Experiments}
\subsection{Experiment Settings}
\textbf{Datasets and Metrics:}
The evaluations are implemented on three benchmark datasets for TIReID, namely CUHK-PEDES~\cite{GNA}, ICFG-PEDES~\cite{SSAN}, and RSTPReid~\cite{DSSL}. \textbf{CUHK-PEDES} comprises 40,206 images and 80,412 descriptions of 13,003 persons. Each image is annotated with 2 descriptions, and the average length of each description is not less than 23 words. Following the dataset split of~\cite{GNA} for equitable comparisons, the training set encompasses 34,054 images and 68,108 descriptions of 11,003 persons, the validation set includes 3,078 images and 6,156 descriptions of 1,000 persons, and the testing set involves 3,074 images and 6,148 descriptions of 1,000 persons. \textbf{ICFG-PEDES} comprises 54,522 image-text pairs of 4,102 persons, with an average description length of 37 words. We adhere to the dataset split strategy outlined in~\cite{SSAN}, utilizing 34,674 image-text pairs of 3,102 persons for model training and reserving the remaining 1,000 persons for evaluation, totaling 19,848 image-text pairs. \textbf{RSTPReid} includes 20,505 images of 4,101 persons, each annotated with 2 descriptions. Each person has 5 images captured by multiple cameras. The average length of each description is not less than 23 words. There are 3,701 persons in the training set, 200 persons in the validation set, and 200 persons in the testing set. 

For performance evaluation, we employ the cumulative matching characteristics curve, a.k.a, Rank-$k$ matching accuracy (R@$k$). This metric signifies the probability that a correct match appears within the top-$k$ ranked retrieved results. Consistent with convention, we report matching accuracy for $k$=${1, 5, 10}$.

\begin{table}[!ht]\small
\setlength{\abovecaptionskip}{0.1cm} 
\setlength{\belowcaptionskip}{-0.2cm}
\centering {\caption{Performance comparison with state-of-the-art methods on CUHK-PEDES. R@1, R@5, and R@10 are listed.}\label{Tab:1}
\renewcommand\arraystretch{1.0}
\begin{tabular}{c|c|c|ccc}
 \hline
  Methods & Pre &Ref & R@1 & R@5 & R@10\\
  \hline  
  





  
  
  


  ACSA~\cite{ACSA} &\multirow{18}{*}{\rotatebox{90}{w/o CLIP}}   & TMM'22  & 63.56   & 81.40  & 87.70 \\
  

  SRCF~\cite{SRCF} &  & ECCV'22 & 64.04   & 82.99   & 88.81 \\

  LBUL~\cite{LBUL} &  & MM'22 & 64.04  & 82.66  & 87.22    \\
  
  
  CAIBC~\cite{caibc} &    & MM'22 & 64.43  & 82.87   & 88.37 \\
  
  AXM-Net~\cite{AXM-Net}  &   & AAAI'22 & 64.44  &80.52  &86.77 \\

  C$_2$A$_2$~\cite{C2A2}  &    & MM'22 & 64.82  & 83.54  & 89.77 \\ 

  LGUR~\cite{lgur}  &    & MM'22  & 65.25  & 83.12  & 89.00 \\

  FedSH~\cite{fedsh} &    & TMM'23 & 60.87 &  80.82 &  87.61 \\
  
  RKT~\cite{RKT} &    & TMM'23 & 61.48  &80.74  &87.28 \\


  PBSL~\cite{pbsl} & & MM'23  & 65.32 & 83.81 & 89.26  \\

  BEAT~\cite{beat} & & MM'23  & 65.61  & 83.45 & 89.57  \\

  MANet~\cite{MANet}  &  & TNNLS'23 & 65.64 & 83.01 & 88.78  \\

  ASAMN~\cite{ASAMN} & & TIP'23  & 65.66 & 84.53 & 90.21   \\

  MSN-BRR~\cite{MSN-BRR} & & ICMR'23  & 65.93 & 83.94 & 90.15  \\

  BDNet~\cite{bdnet} &    & PR'23 & 66.27 & 85.07 & 90.27 \\

  LCR$^2$S~\cite{LCR2S}  &  & MM'23 & 67.36  & 84.19  & 89.62 \\

  TransTPS~\cite{TransTPS} & & TMM'23  & 68.23 & 86.37 & 91.65  \\

  MGCN~\cite{mgcn} & & TMM'23  & 69.40 & 87.07 & 90.82  \\
  \hline

  CFine~\cite{CFine} &\multirow{14}{*}{\rotatebox{90}{w/ CLIP}}   & TIP'23 & 69.57 & 85.93 & 91.15  \\

  Sew~\cite{sew} &    & arXiv'23 & 69.61 & 86.01  & 90.90 \\

  CSKT~\cite{CSKT} &    & arXiv'23 & 69.70 & 86.92 & 91.8 \\

  VLP-TPS~\cite{VLP-TPS}  &  & arXiv'23 & 70.16 & 86.10 & 90.98 \\

  VGSG~\cite{vgsg}  &  & TIP'23 & 71.38 & 86.75 & 91.86 \\

  IRRA~\cite{irra}  &    & CVPR'23 & 73.38 & 89.93 & 93.71 \\

  BiLMa~\cite{BiLMa}  &  & ICCVW'23 & 74.03 & 89.59 & 93.62 \\

  TCB~\cite{TCB}  &  & MM'23 & 74.45 & 90.07 & 94.66  \\
  

  DCEL~\cite{DCEL} &  & MM'23 & 75.02 & 90.89 & 94.52  \\

  SAL~\cite{sal} &  & MMM'24 & 69.14 & 85.90 & 90.81 \\

  EESSO~\cite{eesso} &  & IVC'24 & 69.57 & 85.65 & 90.71 \\
  
  PD~\cite{pd} &  & arXiv'24 & 71.59 & 87.95 & 92.45 \\

  CFAM~\cite{CFAM}  &  & CVPR'24 & 72.87 & 88.61 & 92.87 \\

  TBPS-CLIP~\cite{empirical} &  & AAAI'24 & 73.54 & 88.19 & 92.35 \\
  \hline\hline
  \textbf{Ours} &   & - & \textbf{74.89} & \textbf{89.90} & \textbf{94.17}  \\
  \hline
\end{tabular}}
\end{table}

\textbf{Implementation Details:}
For input images, we uniformly resize them to 384$\times$128 and augment them with random horizontal flipping, random crop with padding, and random erasing. For input texts, the length of the textual token sequence is unified to 77, and we also augment texts by randomly masking out some text tokens with a probability of 15\% and replacing them with the special token $[MASK]$~\cite{bert}. Propot's visual encoder is initialized with the CLIP-ViT-B/6 version of CLIP, where the dimension $d$ of global visual and textual features is set to 512. The Self-Attention Encoder (SAE) and Cross-Attention Decoder (CAD) modules consist of $N_a=1$ block and $N_e=3$ blocks, respectively, each with 8 heads. The length $K$ of the learnable contextual prompt vector in the DPP module is set to 4. In both the SDM and prototype-to-instance contrastive losses, the temperature factor $\tau$ is set to 0.02. The loss balance factors are set to $\lambda_1=0.2$ and $\lambda_2=1.0$. Model training utilizes the Adam optimizer with a weight decay factor of 4e-5. Initial learning rates are set to 1e-5 for the visual/textual encoder and 1e-4 for other network modules. The learning rate is decreased using a cosine learning rate decay strategy, and the training is stopped at 60 epochs. Additionally, the learning rate linearly decays by a factor of 0.1 within the first 10\% of the training epochs for warmup. All experiments are implemented in PyTorch library, and models are trained with a batch size of 64 on a single RTX3090 24GB GPU. 

\subsection{Comparisons with State-of-the-art Models}
In this section, we compare our Propot with current state-of-the-art approaches on all the above three TIReID benchmarks. The methods for comparison are categorized into two sections. The first section includes methods (w/o CLIP) that initialize the backbone solely using single-modal pre-training models (ResNet~\cite{resnet}, ViT~\cite{vit}, BERT~\cite{bert}). In the second section, we present methods (w/ CLIP) that utilize the multi-modal pre-training CLIP~\cite{clip} model to initialize the backbone. Propot falls under the second section. 

\textbf{Evaluation on CUHK-PEDES:}
The performance comparison with state-of-the-art methods on the CUHK-PEDES dataset is summarized in Table \ref{Tab:1}. The proposed Propot framework demonstrates competitive performance at all metrics and outperforms all compared methods except~\cite{DCEL}, achieving remarkable R@1, R@5, and R@10 accuracies of 74.89\%, 89.90\% and 94.17\%, respectively. While our R@1 accuracy is slightly lower (-0.13\%) compared to the optimal method DCEL~\cite{DCEL}, it is essential to note that DECL introduces both mask language modeling and global-local semantic alignment to mine fine-grained matching, resulting in higher computational cost. In contrast, Propot employs only mask language modeling for fine-grained matching. Additionally, as observed in Table~\ref{Tab:4}\#7, even without a local matching module, Propot achieves a noteworthy 74.37\% R@1 accuracy, surpassing most compared methods. This is attributed to the fact that our prototypical prompting can simultaneously model instance-level and identity-level matching. 

\begin{table}[!t]\small
\setlength{\abovecaptionskip}{0.1cm} 
\setlength{\belowcaptionskip}{-0.2cm}
\centering {\caption{Performance comparison with state-of-the-art methods on ICFG-PEDES. R@1, R@5, and R@10 are listed.}\label{Tab:2}
\renewcommand\arraystretch{1.0}
\begin{tabular}{c|c|c|ccc}
\hline
  Methods & Pre & Ref & R@1 & R@5 & R@10 \\
  \hline  


  
  IVT~\cite{ivt} &\multirow{12}{*}{\rotatebox{90}{w/o CLIP}}   & ECCVW'22 & 56.04  & 73.60  & 80.22 \\

  SRCF~\cite{SRCF} & & ECCV'22 & 57.18   & 75.01  & 81.49 \\

  LGUR~\cite{lgur} & & MM'22  & 57.42 & 74.97  & 81.45 \\

  FedSH~\cite{fedsh} &    & TMM'23 & 55.01  & 72.75 & 79.48 \\

  ASAMN~\cite{ASAMN} & & TIP'23  & 57.09 & 76.33 & 82.84   \\


  BDNet~\cite{bdnet} &    & PR'23 & 57.31 & 76.15 & 81.58 \\

  PBSL~\cite{pbsl} & & MM'23  & 57.84 & 75.46 & 82.15  \\

  LCR$^2$S~\cite{LCR2S}  &  & MM'23 & 57.93  & 76.08  & 82.40 \\

  BEAT~\cite{beat} & & MM'23  & 58.25  & 75.92 & 81.96  \\

  MSN-BRR~\cite{MSN-BRR} & & ICMR'23  & 58.63 & 75.78 & 82.15 \\

  MANet~\cite{MANet} &  & TNNLS'23 & 59.44 & 76.80 & 82.75 \\

  MGCN~\cite{mgcn} & & TMM'23  & 60.20 & 76.75 & 83.90  \\
  \hline
  CSKT~\cite{CSKT} & \multirow{12}{*}{\rotatebox{90}{w/ CLIP}}  & arXiv'23 & 58.90 & 77.31 & 83.56 \\
  
  VLP-TPS~\cite{VLP-TPS}  &  & arXiv'23 & 60.64 & 75.97 & 81.76 \\

  CFine~\cite{CFine}  &   & TIP'23 & 60.83 & 76.55 & 82.42  \\

  TCB~\cite{TCB}  &  & MM'23 & 61.60 & 76.33 & 81.90  \\
  
  Sew~\cite{sew}  & & arXiv'23 & 62.29 & 77.15  &  82.52 \\

  VGSG~\cite{vgsg}  &  & TIP'23 & 63.05 & 78.43 & 84.36 \\

  IRRA~\cite{irra}  &    & CVPR'23 & 63.46  & 80.25 &  85.82 \\

  BiLMa~\cite{BiLMa}  &  & ICCVW'23 & 63.83 & 80.15 & 85.74 \\


  DCEL~\cite{DCEL} &  & MM'23 & 64.88 & 81.34 & 86.72  \\

  EESSO~\cite{eesso} &  & IVC'24 & 60.84 & 77.89 & 83.53 \\

  PD~\cite{pd} &  & arXiv'24 & 60.93 & 77.96 & 84.11 \\

  CFAM~\cite{CFAM}  &  & CVPR'24 & 62.17 & 79.57 & 85.32 \\

  SAL~\cite{sal} &  & MMM'24 & 62.77 & 78.64 & 84.21 \\

  TBPS-CLIP~\cite{empirical} &  & AAAI'24 & 65.05 & 80.34 & 85.47 \\
  \hline\hline
  \textbf{Ours} &  & - & \textbf{65.12} & \textbf{81.57}  & \textbf{86.97}   \\
  \hline
\end{tabular}}
\end{table}

\begin{table}[!t]\small
\setlength{\abovecaptionskip}{0.1cm} 
\setlength{\belowcaptionskip}{-0.2cm}
\centering {\caption{Performance comparison with state-of-the-art methods on RSTPReid. R@1, R@5, and R@10 are listed.}\label{Tab:3}
\renewcommand\arraystretch{1.0}
\begin{tabular}{c|c|c|ccc}
 \hline
  Methods & Pre & Ref & R@1 & R@5 & R@10 \\
  \hline



  LBUL~\cite{LBUL} &\multirow{9}{*}{\rotatebox{90}{w/o CLIP}}  & MM'22   & 45.55  & 68.20 & 77.85  \\
  
  IVT~\cite{ivt}  &  & ECCVW'22 & 46.70  & 70.00  & 78.80 \\

  ACSA~\cite{ACSA} &  & TMM'22  & 48.40  & 71.85  & 81.45 \\
  
  C$_2$A$_2$~\cite{C2A2}  &  & MM'22 & 51.55  & 76.75 & 85.15 \\

  PBSL~\cite{pbsl} & & MM'23  & 47.80 & 71.40 & 79.90  \\

  BEAT~\cite{beat} & & MM'23  & 48.10 & 73.10  & 81.30 \\


  MGCN~\cite{mgcn} & & TMM'23  & 52.95 & 75.30 & 84.04  \\

  LCR$^2$S~\cite{LCR2S}  & & MM'23 & 54.95  & 76.65  & 84.70 \\

  TransTPS~\cite{TransTPS} & & TMM'23  & 56.05 & 78.65 & 86.75  \\
  \hline
  CFine~\cite{CFine}  &\multirow{12}{*}{\rotatebox{90}{w/ CLIP}}  & TIP'23 & 50.55 & 72.50 & 81.60  \\
  
  VLP-TPS~\cite{VLP-TPS}  &  & arXiv'23 & 50.65 & 72.45 & 81.20 \\
  
  Sew~\cite{sew} &  & arXiv'23 & 51.95  & 73.50 & 82.45 \\

  CSKT~\cite{CSKT} &   & arXiv'23 & 57.75 & 81.30 & 88.35 \\

  IRRA~\cite{irra}  &    & CVPR'23 & 60.20  &  81.30 &  88.20 \\

  BiLMa~\cite{BiLMa}  &  & ICCVW'23 & 61.20  & 81.50  & 88.80 \\

  DCEL~\cite{DCEL} &  & MM'23 & 61.35 & 83.95 & 90.45  \\


  EESSO~\cite{eesso} &  & IVC'24 & 53.15 & 74.80 & 83.55 \\
  
  PD~\cite{pd} &  & arXiv'24 & 56.65 & 77.40 & 84.70 \\

  CFAM~\cite{CFAM}  &  & CVPR'24 & 59.40 & 81.35 & 88.50 \\

  TBPS-CLIP~\cite{empirical} &  & AAAI'24 & 61.95 & 83.55 & 88.75 \\

  \hline\hline
  \textbf{Ours} &  & - & \textbf{61.87}  & \textbf{83.63}  & \textbf{89.70}   \\
  \hline
\end{tabular}}
\end{table}

\begin{table*}[!ht]\small
\setlength{\abovecaptionskip}{0.1cm} 
\setlength{\belowcaptionskip}{-0.2cm}
\centering {\caption{Ablation study on different components of our Propot on CUHK-PEDES.}\label{Tab:4}
\renewcommand\arraystretch{1.0}
\begin{tabular}  
{m{0.5cm}<{\centering}|m{2.8cm}<{\centering}|m{0.7cm}<{\centering}|m{0.7cm}<{\centering}|m{0.7cm}<{\centering}|m{0.7cm}<{\centering}|m{0.7cm}<{\centering}|m{0.7cm}<{\centering}m{1.0cm}<{\centering}m{1.0cm}<{\centering}|m{1.0cm}<{\centering}m{1.0cm}<{\centering}}
\hline
\hline
\multirow{2}{*}{No.} & \multirow{2}{*}{Methods} & \multirow{2}{*}{IniPt} &\multirow{2}{*}{DPP} & \multicolumn{2}{c|}{IPP}  & \multirow{2}{*}{MLM} & \multirow{2}{*}{R@1} & \multirow{2}{*}{R@5} & \multirow{2}{*}{R@10} & \multirow{2}{*}{Params} & \multirow{2}{*}{FLOPs}\\
\cline{5-6}
          &  &  &  & Intra & Inter  &  &  &       &       &  & \\
\hline
0\# & Baseline  &  &   &  &  &  & 70.06 & 86.91 & 92.01 & 155.26M  & 20.266 \\

2\# & +IniP  &\checkmark  &  &  &    &  & 73.08 & 88.97 & 93.19 & 155.26M  & 20.278  \\

3\# & +DPP  &\checkmark  &\checkmark  &  &    &  &  74.11 & 89.46 & 93.57 & 203.48M  & 25.704  \\

4\# & +IPP (Intra)  &\checkmark &  &\checkmark   &  &  & 73.83 & 89.21 & 93.53  & 158.41M  & 23.058  \\

5\# & +IPP (Inter)  &\checkmark &  &   &\checkmark  &  & 73.65 & 89.42 & 93.69  & 158.41M  & 23.058  \\

6\# & +IPP  &\checkmark &  &\checkmark   &\checkmark  &  & 74.03 & 89.47 & 93.66  & 158.41M  & 25.838  \\

7\# & +DPP+IPP   &\checkmark &\checkmark &\checkmark  &\checkmark  &  & 74.37 & 89.59 & 93.88 & 206.64M  & 31.264  \\

8\# & +DPP+IPP+MLM  &\checkmark &\checkmark  &\checkmark  &\checkmark  &\checkmark & 74.89 & 89.90 & 94.17 & 245.91M  & 37.353  \\
\hline\hline
\end{tabular}}
\end{table*}

\textbf{Evaluation on ICFG-PEDES:}
Table~\ref{Tab:2} reports the comparative results on the ICFG-PEDES dataset. The findings demonstrate that our Propot establishes a new state-of-the-art performance on the ICFG-PEDES dataset, achieving R@1, R@5, and R@10 accuracy scores of 65.12\%, 81.57\%, and 86.97\%, respectively. Notably, Propot significantly outperforms the recent method~\cite{mgcn} based on single-modal pre-training by +4.92\%, +4.82\%, and +3.07\% on R@1, R@5, and R@10, highlighting the substantial impact of multi-modal prior information in the TIReID task. Moreover, Propot surpasses the current state-of-the-art solution TBPS-CLIP~\cite{empirical} by 1.23\% in R@5 and 1.50\% in R@10. The superior performance on ICFG-PEDES confirms the superiority and generalization of our Propot.

\textbf{Evaluation on RSTPReid:}
We finally evaluate the effectiveness of our method on the latest RSTPReid dataset. 
The comparative analysis with state-of-the-art methods is summarized in Table~\ref{Tab:3}. Propot demonstrates commendable performance, achieving competitive results over recent state-of-the-art methods, specifically attaining 61.83\%, 83.45\%, and 89.70\% on R@1, R@5, and R@10. Specifically, our method achieves suboptimal results, with performance slightly lower (-0.08\%) than the optimal method TBPS-CLIP~\cite{empirical}. TBPS-CLIP represents an empirical study incorporating CLIP into the TIReID task, utilizing various data augmentation and training tricks. In contrast, our approach uses only basic data augmentation without additional tricks. Propot still achieves superior performance on the RSTPReid dataset with complex scenes, underscoring the efficacy of our prototype learning method in seamlessly integrating diverse identity information for modeling identity-level matching. This further establishes the generalization and robustness of Propot.

\subsection{Ablation Studies}
We conduct ablation experiments for Propot on CUHK-PEDES using the default settings above. 
In addition, Baseline solely includes visual and textual encoders initialized by CLIP, which is trained using SDM and cross-entropy loss, with the training settings aligned with those of Propot.

\textbf{Contributions of Proposed Components:}
In Table~\ref{Tab:4}, we assess the contribution of each module of Propot, including the initial prototype (IniPt) generated by CLIP, the DPP module, the IPP module, and the MLM module. Baseline achieves a notable R@1 accuracy of 70.06\% due to the rich multi-modal prior information of CLIP. The introduction of our prototype learning process based on Baseline, aimed at modeling identity-level matching, yields several key observations. Firstly, utilizing only the initial prototype generated by CLIP to supervise identity-level matching leads to significant performance improvements (+3.02\% R@1 improvement over Baseline), affirming the feasibility of our approach and the importance of identity-level matching. Secondly, the incorporation of the DPP module to update the initial prototype results in a 4.05\% R@1 accuracy improvement over Baseline, demonstrating the effective adaptation of the initial prototype to the TIReID task. Thirdly, when the IPP module is employed to update the prototype, whether conditioned on intra-modal or inter-modal instances, substantial performance enhancements are observed (+3.77\% or +3.59\% R@1 improvement over Baseline). The performance is further elevated when both IPP modules are utilized concurrently, underscoring the IPP module's capacity to enrich the prototype with instance information, ensuring its diversity. Fourthly, the collaborative use of DPP and IPP modules further enhances the R@1 accuracy to 74.37\%, surpassing most state-of-the-art methods in Table~\ref{Tab:1}. This is attributed to our Propot's ability to model instance-level and identity-level matching simultaneously. Finally, the incorporation of the local matching module culminates in the best performance for Propot, achieving a remarkable 74.89\% R@1.

\begin{figure}[!t]
\setlength{\abovecaptionskip}{0.1cm} 
\setlength{\belowcaptionskip}{-0.2cm}
\subfigbottomskip=-2pt
\subfigcapskip=-2pt
\centering
\subfigure[impact of $K$] {\includegraphics[height=1.2in,width=1.65in,angle=0]{images/ctx_num.pdf}}
\subfigure[impact of $N_a$] {\includegraphics[height=1.2in,width=1.65in,angle=0]{images/lptlayers.pdf}}
\subfigure[impact of $N_e$] {\includegraphics[height=1.2in,width=1.65in,angle=0]{images/isplayers.pdf}}
\subfigure[impact of $\lambda_1$] {\includegraphics[height=1.2in,width=1.65in,angle=0]{images/plossw.pdf}}
\caption{Effects of four hyper-parameters on CUHK-PEDES, including contextual vector length $K$, the block number $N_a, N_e$, and loss weight $\lambda_1$.}
\label{Fig:3}
\end{figure}

\textbf{Impact of contextual vector length $K$ in DPP:}
To facilitate the adaptation of the initial prototype to the TIReID task, we introduce a set of learnable contextual prompt vectors for each prototype in the DPP module. The length $K$ of this set of vectors is a crucial parameter influencing the prototype's adaptation. To investigate the impact of different vector lengths, we conduct several experiments by varying the value of the context vector length from 1 to 6 and report the results in Figure~\ref{Fig:3} (a). The observed trend indicates that larger values of vector length result in superior performance, as they provide sufficient contextual parameters to extract TIReID task information. Notably, when $K$ equals 4, the performance reaches a peak of 74.89\%. However, excessively large vector lengths may introduce information redundancy, leading to overfitting and a surge in computational costs. Therefore, we set $K$ to 4 in our experiments to strike a balance between performance and efficiency.

\textbf{Influence of $N_a, N_e$:}
The parameters $N_a$ and $N_e$ correspond to the number of blocks of SAE and CAD in DPP and IPP modules, respectively, and their impact on performance is shown in Figure~\ref{Fig:3} (b) and (c). Regarding $N_a$, it is evident that when its value exceeds 3, the performance experiences a significant drop. This suggests that an excessive number of SAE parameters might hinder the learning of contextual prompt vectors. Therefore, we set $N_a$ to 1 in our experiments, which yields the optimal result. In comparison, our model demonstrates less sensitivity to $N_e$, and its curve exhibits a relatively flat trend. Nevertheless, a $N_e$ that is excessively large would introduce a surplus of parameters, leading to a considerable increase in computational costs. Consequently, we set $N_e$ to 3 in our experiments, achieving superior performance while maintaining efficiency.

\begin{table}[!t]\small
\setlength{\abovecaptionskip}{0.1cm} 
\setlength{\belowcaptionskip}{-0.2cm}
\centering {\caption{Ablation study of prototype aggregation schemes on CUHK-PEDES.}\label{Tab:5}
\renewcommand\arraystretch{1.0}
\begin{tabular}{m{3cm}<{\centering}|m{1.2cm}<{\centering}m{1.2cm}<{\centering}m{1.2cm}<{\centering}}
 \hline
  Method  & Rank-1 & Rank-5 & Rank-10 \\
  \hline
  Sum      & 74.30  &  89.86 & 93.84   \\
  
  Average   & 74.29 & 89.39 & 93.34  \\

  MLP     & 73.51 & 89.56 & 93.76 \\

  Parameter   & 74.19 & 89.10 & 93.18  \\
  \hline\hline
  APA (Ours)   & 74.89 & 89.90 & 94.17  \\
  \hline
\end{tabular}}
\end{table}

\textbf{Impact of weight $\lambda_1$ in the objective function:}
The parameter $\lambda_1$ represents the weight of the prototype-to-instance contrastive loss, governing the strength of identity-level matching. To assess the impact of this parameter, we conduct several experiments, varying the value of $\lambda_1$ from 0.01 to 10, with results depicted in Figure~\ref{Fig:3} (d). It is evident from the results that as $\lambda_1$ increases, the performance gradually improves, reaching its peak when $\lambda_1$ is set to 0.2. Subsequently, performance begins to decline as $\lambda_1$ continues to increase, ultimately collapsing when $\lambda_1$ equals 10. This behavior is attributed to the fact that a very small $\lambda_1$ fails to ensure the effective diffusion of identity information to each instance, rendering identity-level matching ineffective. Conversely, an excessively large $\lambda_1$ results in the over-diffusion of identity information, disrupting instance-level matching. Therefore, we set $\lambda_1$ to 0.2 to strike a balance between instance-level and identity-level matching.

\begin{figure*}[!t]
  \centering
  \setlength{\abovecaptionskip}{0.1cm} 
  \setlength{\belowcaptionskip}{-0.2cm}
  \includegraphics[width=\linewidth]{images/retrieval-result.pdf}\\
  \caption{Retrieval result comparisons of Baseline and our Propot on CUHK-PEDES. The matched and mismatched person images are marked with green and red rectangles, respectively.}
  \label{Fig:4}
\end{figure*}

\textbf{Different Prototype Aggregation Schemes:}
To adaptively aggregate multiple prototypes generated by different modules, we devised an Adaptive Prototype Aggregation (APA) module. To demonstrate the superiority of the APA module, we conducted a comparison with several common aggregation schemes, including (1) summing multiple prototypes; (2) averaging multiple prototypes; (3) employing a multi-layer perceptron (MLP) network to generate individual weights for each prototype; (4) parameterizing the weights of each prototype to learn simultaneously with the network. The comparison results are summarized in Table~\ref{Tab:5}. Analysis of the table leads to the following conclusions. Aggregation schemes that introduce parameters to learn aggregation weights perform less favorably than non-parametric aggregation schemes. This discrepancy arises because the introduction of additional parameters amplifies the optimization difficulty and uncertainty of the network. Furthermore, superior performance can be achieved with simple summation or average aggregation schemes, underscoring the feasibility and effectiveness of our idea. The excellence of our aggregation scheme is attributed to its utilization of the initial prototype generated by CLIP as a baseline for adaptive aggregation of other prototypes. The quality of the initial prototype is assured by the powerful semantic understanding capabilities of CLIP.

\textbf{Qualitative Results:}
To further highlight the effectiveness of the proposed Propot, we present some representative retrieval results in Figure~\ref{Fig:4}. For each text description in the query set, Figure 4 displays the top 10 gallery images retrieved by both the baseline and our Propot. The green and red rectangles denote correct and incorrect retrieval results for the given query text, respectively. Baseline solely focuses on instance-level matching between modalities, while our Propot incorporates prototype prompting based on Baseline to model identity-level matching. The examples illustrate that our Propot excels in handling challenging scenarios where the baseline struggles, ensuring that pedestrian images with the same identity as the given query text are ranked highly. This visually underscores the significance of identity-level matching, with our method demonstrating superior performance by concurrently modeling both instance-level and identity-level matching.

\section{Conclusion}
In this study, to model identity-level matching for TIReID, we present Propot, a conceptually simple framework for identity-enriched prototype learning. The framework follows the ‘initialize, adapt, enrich, then aggregate’ pipeline. Initially, we generate robust initial prototypes leveraging the capabilities of CLIP. Subsequently, the Domain-conditional Prototypical Prompting (DPP) module is introduced to prompt and adapt the initial prototypes to the specificities of the TIReID task. To ensure prototype diversity, the Instance-conditional Prototypical Prompting (DPP) is devised, enriching the prototypes through both intra-modal and inter-modal instances. Lastly, an adaptive prototype aggregation module is employed to effectively aggregate multiple prototypes, followed by the diffusion of their rich identity information to instances, thereby facilitating the modeling of identity-level matching. Through extensive ablations and comparative experiments conducted on three widely used benchmarks, we demonstrate the superiority and effectiveness of the proposed Propot framework.

\begin{acks}
This work was supported by...
\end{acks}

\clearpage
\bibliographystyle{ACM-Reference-Format}
\bibliography{sample-base}
